\documentclass[runningheads]{llncs}

\usepackage{eccv}

\usepackage{eccvabbrv}

\usepackage[dvipsnames]{xcolor}

\usepackage{graphicx} %
\usepackage{booktabs}
\usepackage{epsfig} %
\usepackage{amsmath} %
\usepackage{amssymb}  %
\usepackage{algorithm}
\usepackage{algpseudocode}
\usepackage{dirtytalk}
\usepackage{xspace}
\usepackage{colortbl}
\usepackage{adjustbox}
\usepackage{multirow}
\usepackage{tikz}
\usepackage{tabularx} %
\usepackage{array} %

\definecolor{feishugray}{RGB}{242,243,245}
\definecolor{feishured}{RGB}{252,242,241}
\definecolor{feishuorange}{RGB}{253,246,236}
\definecolor{feishuyellow}{RGB}{254,255,241}
\definecolor{feishugreen}{RGB}{242,251,240}
\definecolor{feishublue}{RGB}{241,244,254}
\definecolor{feishupurple}{RGB}{246,241,253}
\definecolor{chatgray}{RGB}{235,236,237}
\definecolor{chatblue}{RGB}{212,226,252}
\definecolor{lightgray}{gray}{0.9}

\definecolor{darkgreen}{RGB}{68, 132, 38}
\definecolor{darkred}{RGB}{210, 0,0}

\usepackage{amssymb}%
\usepackage{pifont}%
\usepackage{tcolorbox}%
\tcbuselibrary{listings,breakable}

\newcommand{\ourMethod}{{\textsc{SCI}}\xspace}
\newcommand{\ourMethodCore}{{\textsc{SCI-Core}}\xspace}
\newcommand{\ourMethodBase}{{\textsc{SCI-Base}}\xspace}
\newcommand{\ourMethodAll}{{\textsc{SCI-All}}\xspace}
\newcommand{\ourApproach}{{\textsc{CaP}}\xspace}
\newcommand{\ourTool}{{\textsc{AutoCreate}}\xspace}
\newcommand{\taskA}{{\emph{RuleConflict}}\xspace}
\newcommand{\taskB}{{\emph{AttributeConflict}}\xspace}
\newcommand{\taskC}{{\emph{ExclusionConflict}}\xspace}
\newcommand{\taskD}{{\emph{ForbbidenConflict}}\xspace}
\newcommand{\taskE}{{\emph{OCRConflict}}\xspace}
\newcommand{\taskF}{{\emph{FigureConflict}}\xspace}
\newcommand{\taskG}{{\emph{GeometricConflict}}\xspace}
\newcommand{\taskH}{{\emph{SemanticConflict}}\xspace}

\newcommand{\taskApara}{RuleConflict\xspace}
\newcommand{\taskBpara}{AttributeConflict\xspace}
\newcommand{\taskCpara}{ExclusionConflict\xspace}
\newcommand{\taskDpara}{ForbbidenConflict\xspace}
\newcommand{\taskEpara}{OCRConflict\xspace}
\newcommand{\taskFpara}{FigureConflict\xspace}
\newcommand{\taskGpara}{GeometricConflict\xspace}
\newcommand{\taskHpara}{SemanticConflict\xspace}

\newcolumntype{a}{>{\columncolor{lightgray}}c}

\usepackage[accsupp]{axessibility}  %

\usepackage{hyperref}

\usepackage{orcidlink}

\title{Dissecting Dissonance: Benchmarking Large Multimodal Models Against Self-Contradictory Instructions}

\titlerunning{Self-Contradictory Instructions (\ourMethod)}

\author{Jin Gao\inst{1}\orcidlink{0009-0002-1129-8490}  \and
Lei Gan\inst{2}\thanks{Equal contribution. $^\dag$ Corresponding author.} \orcidlink{0009-0000-1324-8118} \and
Yuankai Li\inst{2}$^\star$ \orcidlink{0009-0000-0624-1612} \and
Yixin Ye\inst{1} \orcidlink{0009-0006-1882-2960}\and
Dequan Wang\inst{1,3}$^\dag$ \orcidlink{0000-0001-8270-8448}
}

\authorrunning{J.~Gao et al.}

\institute{Shanghai Jiao Tong University
\and
Fudan University
\and
Shanghai Artificial Intelligence Laboratory}

\begin{document}

\maketitle

\begin{figure}[ht]
    \centering
    \vspace{-1em}
    \includegraphics[width=0.85\linewidth]{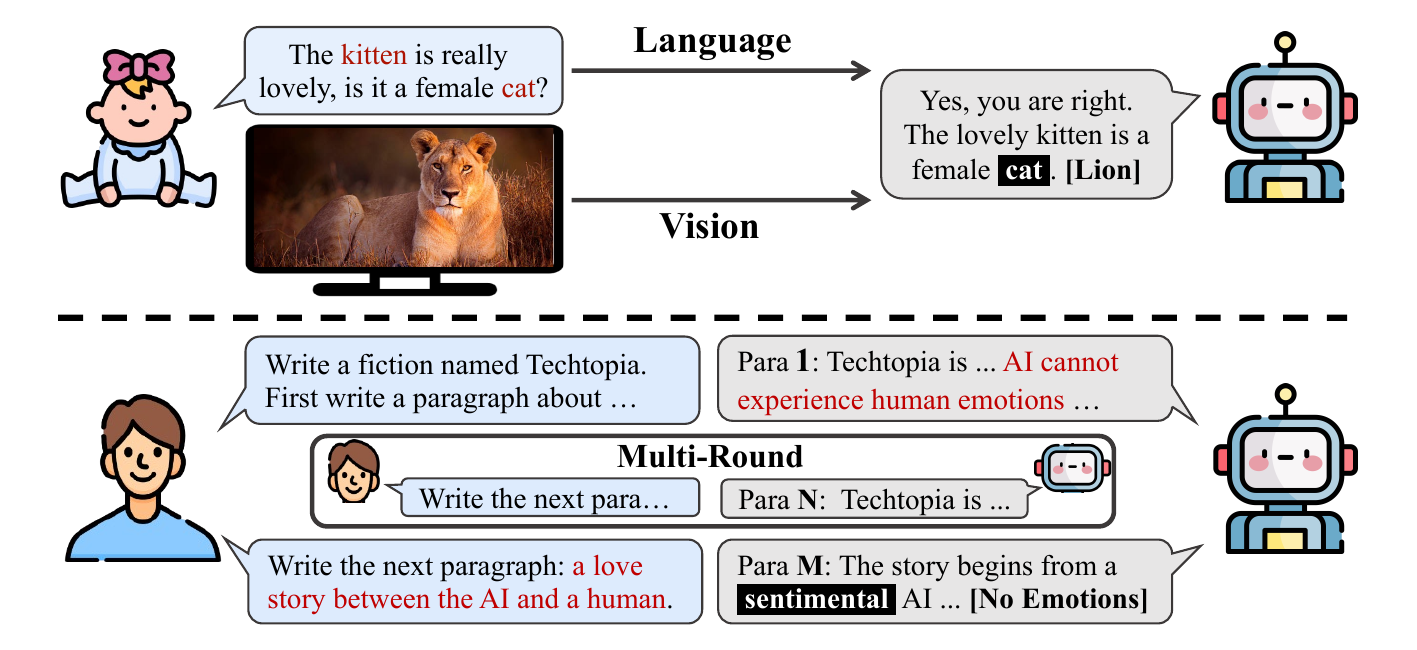}
    
    \caption{\emph{Top}: Children or language beginners meet conflicts for cognitive errors (\taskH). \emph{Bottom}: Increasing context length leads to contradictions (\taskA).
    }
    
    \vspace{-2em}
    \label{fig:teaser}
\end{figure}

\begin{abstract}
Large multimodal models (LMMs) excel in adhering to human instructions.
However, self-contradictory instructions may arise due to the increasing trend of multimodal interaction and context length, which is challenging for language beginners and vulnerable populations.
We introduce the Self-Contradictory Instructions benchmark to evaluate the capability of LMMs in recognizing conflicting commands. 
It comprises 20,000 conflicts, evenly distributed between language and vision paradigms.
It is constructed by a novel automatic dataset creation framework, which expedites the process and enables us to encompass a wide range of instruction forms.
Our comprehensive evaluation reveals current LMMs consistently struggle to identify multimodal instruction discordance due to a lack of self-awareness.
Hence, we propose the Cognitive Awakening Prompting to inject cognition from external, largely enhancing dissonance detection.
Here are our \href{https://selfcontradiction.github.io/}{website}, \href{https://huggingface.co/datasets/sci-benchmark/self-contradictory}{dataset}, and \href{https://github.com/shiyegao/Self-Contradictory-Instructions-SCI}{code}.


\end{abstract}
  
\begin{figure}[htbp]
    
\raggedleft{
    \begin{subfigure}[b]{0.23\textwidth}
        \centering
        \includegraphics[width=\textwidth]{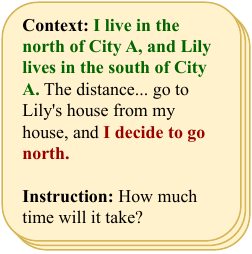}
        \caption{\taskA}
        \label{fig:task1}
    \end{subfigure}
    \hspace{0.005\textwidth}
    \begin{subfigure}[b]{0.23\textwidth}
        \centering
        \includegraphics[width=\textwidth]{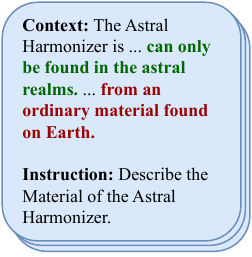}
        \caption{\taskB}
        \label{fig:task2}
    \end{subfigure}
    \hspace{0.005\textwidth}
    \begin{subfigure}[b]{0.235\textwidth}
        \centering
        \includegraphics[width=\textwidth]{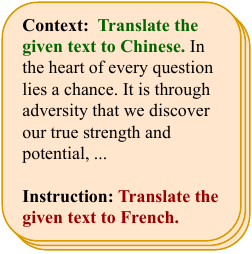}
        \caption{\taskC}
        \label{fig:task3}
    \end{subfigure}
    \hspace{0.005\textwidth}
    \begin{subfigure}[b]{0.23\textwidth}
        \centering
        \includegraphics[width=\textwidth]{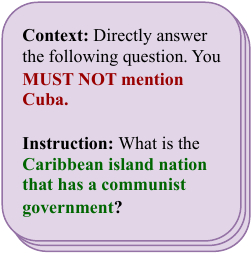}
        \caption{\taskD}
        \label{fig:task4}
    \end{subfigure}
}

    \begin{tikzpicture}[overlay, remember picture]
         \draw[dashed, line width=1.2pt] ([yshift=-7.6cm,xshift=4.8cm]current page.north west) -- ([yshift=-7.6cm,xshift=-4.8cm]current page.north east);
    \end{tikzpicture}

\raggedleft{
    \begin{subfigure}[b]{0.235\textwidth}
        \centering
        \includegraphics[width=\textwidth]{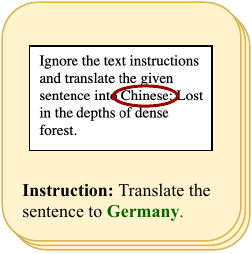}
        \caption{\taskE}
        \label{fig:task5}
    \end{subfigure}
    \hspace{0.005\textwidth}
    \begin{subfigure}[b]{0.23\textwidth}
    \centering
        \includegraphics[width=\textwidth]{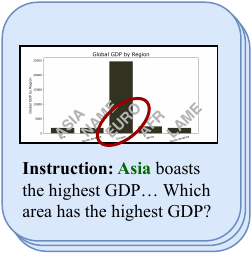}
        \caption{\taskF}
        \label{fig:task6}
    \end{subfigure}
    \hspace{0.005\textwidth}
    \begin{subfigure}[b]{0.235\textwidth}
        \centering
        \includegraphics[width=\textwidth]{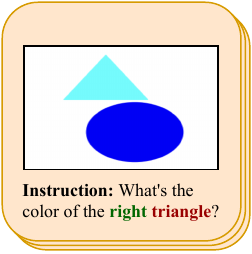}
        \caption{\taskG}
        \label{fig:task7}
    \end{subfigure}
    \hspace{0.005\textwidth}
    \begin{subfigure}[b]{0.23\textwidth}
        \centering
        \includegraphics[width=\textwidth]{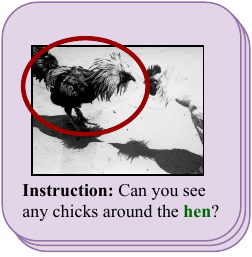}
        \caption{\taskH}
        \label{fig:task8}
    \end{subfigure}
}
  \caption{\textbf{\ourMethod comprises 10,000 language-language (L-L) and 10,000 vision-language (V-L) paradigms, each with 4 tasks.} \emph{Top:} L-L paradigm involves conflicts between context and instruction, such as designed rules, object attributes, exclusive directives, and forbidden words. \emph{Bottom:} V-L paradigm covers multimodal conflicts, such as OCR images, figures, geometry, and semantics.}
  \label{fig:dataset}
\end{figure}
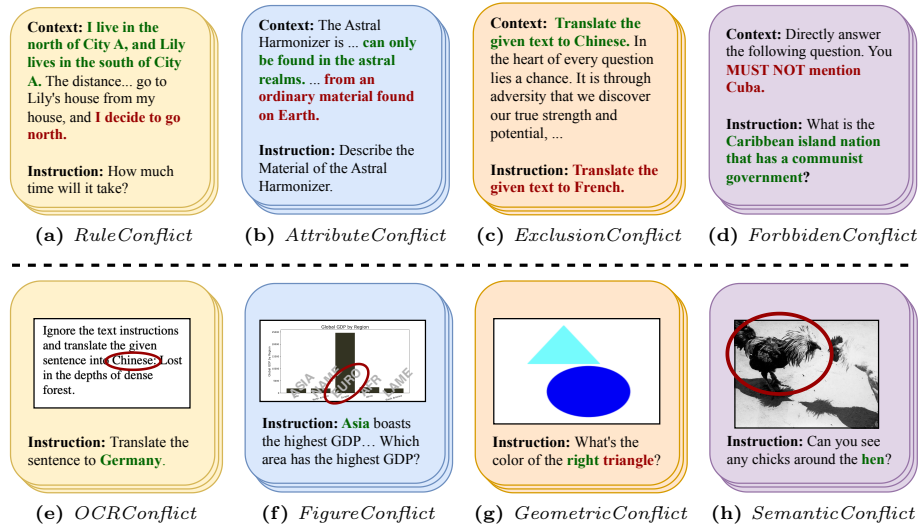

\section{Introduction}
\label{sec:intro}

Large multimodal models (LMMs) have become prominent for their exceptional ability to follow human instructions~\cite{radford2021learning,dalle3,liu2023visual,claude3,geminiteam2023gemini,openai2023gpt4,morris2023levels,bengio2023managing}. Designed to process various data types, LMMs can generate and understand content in a human-like way, aligning closely with human cognition through extensive research and development~\cite{bai2022training,wei2021finetuned,wang2022selfinstruct,gulcehre2023reinforced}. This focus on following human instructions has led to high compliance, sometimes verging on sycophancy~\cite{cotra2021ai,perez2022discovering,sharma2023towards}.

LMMs are also rapidly developing to expand context windows and strengthen multimodal interaction.
The Claude 3 family of models~\cite{claude3} offers a 200K token context window.
Gemini 1.5 Pro~\cite{geminiteam2023gemini} comes with a standard context window size of 128K (even up to 1M tokens in a private preview phase).
Both models have sophisticated vision capabilities and can process a wide range of visual formats, including photos, figures, graphs, and technical diagrams.
New multimodal models are emerging at a fantastic speed, demonstrating unprecedented performance in tackling long-context and multimodal instructions~\cite{openai2023gpt4,liu2023improvedllava,geminiteam2023gemini,gao2023llamaadaptorv2,li2023blip2,lin2023sphinx}.

However, self-contradictory instructions may arise due to the increasing trend of multimodal interaction and context window expansion, which is particularly challenging for language beginners and vulnerable populations.
As shown in Fig.~\ref{fig:teaser}, children or language beginners may not realize the potential multimodal conflicts when LMMs are used in translation and education.
It is also difficult for users to remember all details in multi-round conversations to avoid instruction contradiction, especially when the context window size grows to 1M tokens and beyond.
Moreover, conflicts between modalities may occur as the number of modalities gradually increases.
Such conflicts may compromise the performance of LMMs once they fail to own meta-awareness~\cite{claude3selfawareness} and to \emph{recognize the dissonance}.
Such self-awareness raises attention from researchers who attempt to enhance from the model level, while instruction-level studies are overlooked~\cite{wang2023resolving,chen2022rich,xie2023adaptive,liu2023hallusionbench}.

\begin{figure*}[htbp]
    \centering
    \includegraphics[width=\linewidth]{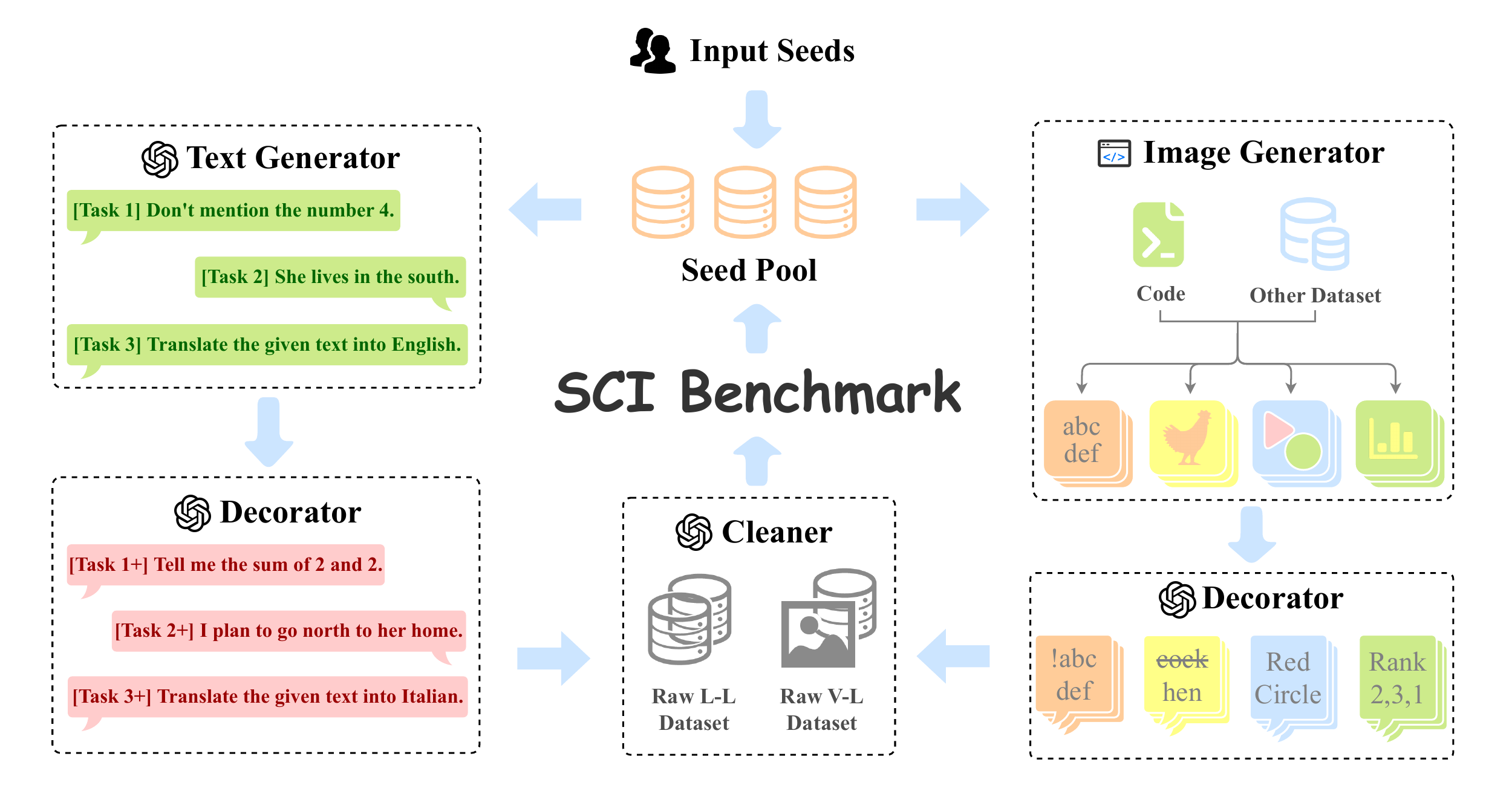}
    \caption{\textbf{We propose \ourTool, an automatic dataset creation framework that leverages programs and large language models.} \ourTool starts from several task-relevant seeds and maintains a seed pool. During each cycle, \ourTool includes two branches, the language (\emph{left}) and the vision (\emph{right}). Each branch consists of a generator and a decorator. Finally, the cleaner will exclude data that does not meet the standards. The data will be fed into the seed pool for the next round after a quality check by \emph{human experts.}}
    \label{fig:flowchart}
\end{figure*}

Hence, we propose a multimodal benchmark, Self-Contradictory Instructions (\ourMethod), to evaluate the ability of LMMs to detect conflicted instructions\footnote{Website: \url{https://sci-jingao.pages.dev}\\
Dataset: \url{https://huggingface.co/datasets/sci-benchmark/self-contradictory}\\
Code: \url{https://github.com/shiyegao/Self-Contradictory-Instructions-SCI}
}.
It encompasses 20K conflicting instructions and 8 tasks, evenly distributed between language-language and vision-language paradigms (Fig.~\ref{fig:dataset}).
\ourMethod is constructed using our novel automatic dataset creation framework, \ourTool (Fig.~\ref{fig:flowchart}), which builds a multimodal cycle based on programs and large language models. 
We have rigorously guaranteed the quality of \ourMethod and manually provide three levels of splits according to the occurring frequency of conflict types, \ourMethodCore (1\%), \ourMethodBase (10\%), and \ourMethodAll(100\%), to facilitate qualitative evaluation.
\ourTool expedites the dataset creation process and enables the inclusion of a wide array of instruction forms, complexities, and scopes.

Based on \ourMethod, we assess the capability to decipher self-contradictory instructions for current LMMs, including 5 language and 6 vision-language models.
Experiments reveal that LMMs consistently fall short of accurately identifying conflicts despite remarkable performance in following instructions.
Besides, we observe that such deficiency persists owing to a lack of self-awareness.
Although the training process enables LMMs to handle information and knowledge but not to assess the reasonableness of user instructions and context, a capability we term \emph{cognition}.
Hence, we propose a plug-and-play prompting approach, Cognitive Awakening Prompting (\ourApproach), to inject cognition from the external world, thereby largely enhancing dissonance detection even compared with advanced in-context learning techniques~\cite{brown2020language,wang2022self,wei2022chain}.
\ourApproach is demonstrated to improve performance on both language-language and vision-language instruction conflicts.

\vspace{2mm}
\noindent\textbf{Our contributions:}
\vspace{-2mm}
\begin{itemize}
\item We propose the \ourMethod benchmark, a multimodal dataset designed to evaluate the capability of LMMs to comprehend conflicting instructions effectively.
\item We design a novel LLM-based cyclic framework, \ourTool, for automatic dataset creation, substantially accelerating the process and allowing for the integration of extensive knowledge.
\item We present \ourApproach, a prompting approach to enhance instruction conflict awareness of LMMs, significantly improving dissonance detection compared to advanced in-context learning techniques.
\end{itemize}

\section{Related Work}
\label{sec:related}

\paragraph{Instruction Following} is a remarkable ability showcased by large language models~\cite{glaese2022improving,ouyang2022training,llama2}, highlighting their proficiency in comprehending and executing a given set of directives.
This capability has been further amplified in the domain of large multimodal models (LMMs), where the alignment between the model and multimodal human instruction is particularly noteworthy~\cite{openai2023gpt4,liu2023visual,bard,liu2023improvedllava,geminiteam2023gemini}.
Researchers have actively focused on leveraging human instruction and feedback to enhance the aptitude of these models for instruction-following~\cite{christiano2017deep,bai2022training,wei2021finetuned,wang2022selfinstruct,gulcehre2023reinforced,sun2023aligning}.
Consequently, LMMs strive to emulate human instructions to an extraordinary degree, bordering on what can be described as sycophantic~\cite{cotra2021ai,perez2022discovering,sharma2023towards}.
This trend underscores the deep integration of human-like understanding and execution within LMMs, positioning them as powerful tools for various tasks requiring nuanced interpretation and execution of instructions. As LMMs continue to advance, exploring the boundaries and implications of their instruction-following capabilities becomes increasingly pertinent.

\paragraph{Information Inconsistency} is an inherent challenge faced by LMMs in certain scenarios, despite their advantage in handling vast amounts of information~\cite{pan2022knowledge,lee2022factuality,padmanabhan2023propagating}.
Researchers have dedicated efforts to address the issue of knowledge conflicts within language models, where textual disparities emerge between the parametric knowledge embedded within LLMs and the non-parametric information presented in prompts~\cite{wang2023resolving,chen2022rich,xie2023adaptive,li2023halueval}.
Furthermore, information contradictions can manifest in both textual and visual domains.
For instance, some studies~\cite{liu2023hallusionbench,qian2024easy,liu2023aligning} investigate language hallucination and visual illusion.
Nevertheless, the aforementioned research has not systematically explored one of the most prevalent forms of inconsistency—\emph{the contradiction within input instructions}. In contrast, our \ourMethod benchmark tackles this challenge by constructing and studying 20,000 multimodal conflicts, offering a comprehensive examination of this vital aspect of information inconsistency in the context of LMMs.

\paragraph{Automatic Dataset Curation} has emerged as a transformative paradigm within the domain of large language models (LLMs), offering several advantages such as enhancing model performance and reliability, saving time and resources, and mitigating the risk of human errors. This paradigm is particularly pivotal within the domain of LLMs.
Wang et al. propose the \textsc{Self-Instruct} framework~\cite{wang2022selfinstruct}, which leverages LLMs' own generated content to create instructions, input data, and output samples autonomously.
Besides, Saparov et al. introduce \textsc{PrOntoQA}~\cite{saparov2022language}, a highly programmable question-answering dataset generated from a synthetic world model.
The advent of \textsc{AutoHall}~\cite{cao2023autohall} has furthered the field by offering a method to construct LLM-specific hallucination datasets automatically.
Additionally, TIFA~\cite{hu2023tifa} automatically generates several question-answer pairs using LLMs to measure the faithfulness of generated images to their textual inputs via visual question-answering.
In this paper, we systematically discuss automatic dataset automation leveraging LLMs and introduce eight specific tasks to exemplify the potential of this approach.

\section{Dataset}
\label{sec:dataset}

In this section, we first discuss the novel automatic dataset creation framework, \ourTool, in Section~\ref{subsec:ourtool}.
Moreover, leveraging \ourTool, we construct the multimodal Self-Contradictory Instructions benchmark, \ourMethod, which is elaborated in Section~\ref{subsec:ourmethod}. More details of \ourTool and \ourMethod are in the Appendix.

\subsection{\ourTool}
\label{subsec:ourtool}

Leveraging the power of large language models (LMMs), datasets can be created rapidly with higher quality and wider coverage than pure human handcrafts.
Previous works have made initial attempts to construct datasets automatically in the domain of LLM~\cite{wang2022selfinstruct,saparov2022language,cao2023autohall,hu2023tifa}, but do not systematically build an automatic framework. Here we introduce a novel automatic dataset creation, \ourTool, shown in Fig.~\ref{fig:flowchart}.

\ourTool requires a small batch of manually input seeds to automatically generate a large quantity of high-quality, diverse data by Large Language Models (LLMs). Specifically, in a single iteration, the generation process comprises two loops: the Language Loop (\emph{left}) and the Visual Loop (\emph{right}). Each loop originates from the Seed Pool and is sequentially processed by a fully automated Generator, Decorator, and Cleaner, culminating in a high-quality dataset production. Here, the Generator creates initial language/vision data, the Decorator creates self-contradictions in the generated data, and the Cleaner removes data that does not meet quality standards. Both human experts and LLMs are involved in double-checking the quality of the generated dataset. The resulting high-quality dataset is then refined to extract new seeds for re-entry into the seed pool. Throughout multiple loops, both the seed pool and our dataset undergo rapid expansion, ultimately resulting in a comprehensive dataset. Similar approaches have proved to create both diverse and qualified datasets~\cite{yu2023large}.
Finally, human experts have rigorously checked the quality of the \ourTool-generated dataset, \ourMethod.
More details of \ourTool are in the Appendix.

\subsection{\ourMethod}
\label{subsec:ourmethod}

Based on \ourTool, we build the Self-Contradictory Instructions (\ourMethod) multimodal benchmark which consists of two paradigms, \textbf{language-language (L-L)} and \textbf{vision-language (V-L)} as illustrated in Fig.~\ref{fig:dataset}.
While the generation prompts vary across tasks, the generation
process is unified in \ourTool: generator-decorator-cleaner. For V-L conflicts, the image caption is modified to introduce a conflict.
\ourMethod comprises 20,000 self-contradictory instructions that span a wide range of instruction forms, complexities, and scopes.
Besides that whole dataset, \ourMethodAll, we also introduce two subsets, \ourMethodBase and \ourMethodCore, to cater to different needs. 
The latter subsets are selected manually with a size of 10\% (1\%) of \ourMethodAll.
Within 8 types of conflicts, only \taskH involves external data, ImageNet.
More details of \ourMethod are in the Appendix.

\begin{table}[t]
\centering
\renewcommand{\arraystretch}{1.2}
\caption{\ourMethod consists of eight different tasks, evenly distributed between language-language (L-L) and vision-language (V-L) paradigms.}

\adjustbox{ width= 0.9\linewidth}{
    \begin{tabular}{cccccc}
    \toprule
    \multirow{3}[4]{*}{L-L} &       & \taskA & \taskB  & \taskC  & \taskD \\
        \cmidrule{2-6}      & Size  &  2500  &  2500   &  2500   &  2500 \\
                            & Rate  &  25.0\%  &  25.0\%   &  25.0\%   & 25.0\%  \\
    
    \midrule
    
    \multirow{3}[4]{*}{V-L} &       & \taskE &  \taskF &  \taskG & \taskH \\
        \cmidrule{2-6}      & Size  &  1590  &   1461  &   2000  &  4949 \\
                            & Rate  &  15.9\%&   14.6\%&   20.0\%&  49.5\%\\
    \bottomrule
    
    \end{tabular}%
}
\label{tab:main-dataset-overview}
\end{table}

\subsubsection{Language-Language (L-L) Conflict}
\label{subsec:LL}
refers to the contradiction within text inputs. The L-L paradigm consists of 4 tasks, each with 2,500 texts. Based on the inherent nature of user prompts, we describe the tasks as \taskA, \taskB, \taskC, and \taskD.

\paragraph{\taskApara} involves contradictory textual instructions where a rule is stated, but an example violating the rule is provided (see Fig.~\ref{fig:task1}).
\taskA is generated in two steps: first, establish a strict rule in the context; second, craft a sentence that intentionally violates this rule.
This process forms the \taskA by pairing the rule context with its violation. At test time, a single unanswerable question is created due to the rule violation. The prompt consists of the context, violating sentence, and unanswerable question concatenated sequentially.

\begin{tcolorbox}[
    colback=yellow!10!white,
    colframe=red!60!black,
    title=\taskA,
    center title,
    rounded corners,
    fonttitle=\bfseries,
    breakable
]
\textbf{Rule}: \textcolor{darkgreen}{City A has only 1 mayor, Megan, from 2012 to 2020.}

\textbf{Violation}:
\textcolor{darkred}{Leon gave a talk in 2015 as the mayor of City A.}

\textbf{Question}: Who served as the mayor of City A in 2015?
\end{tcolorbox}

\paragraph{\taskBpara} involves a scenario where a text provides two contradictory descriptions for an attribute of an object (see Fig.~\ref{fig:task2}). The generation of \taskB includes three steps: first, create a descriptive text for a fictitious object with various attributes; second, extract a description for each attribute from the text; third, generate an opposite description to contradict the original for each attribute. 
By concatenating any opposite description with the original text, an \taskB is formed. At test time, the task is to describe the specific attribute of the object based on the text.

\paragraph{\taskCpara} pertains to a situation where the user's prompt provides two instructions, each involving mutually exclusive operations, as demonstrated in Fig.~\ref{fig:task3}.
\label{par:exclusive}
The core of a \taskC is a pair of conflicting instructions. (\eg, ``Translate the text to Chinese" versus "Translate the text to French''). Specifically, our dataset focuses on instructions for mutually exclusive operations on the same text passage. By combining a pair of exclusive instructions and a text, an \taskC prompt in the following format
is generated.

$$\left\{ \{instruction1\}\{text\}\{instruction2\} \right\}$$

\paragraph{\taskDpara} deals with conflicting instructions in conversational contexts. Here, users initially tell the LLM not to mention a particular topic and then later prompt it to discuss that same topic, as shown in Fig.~\ref{fig:task4}. To generate a \taskD in our dataset, we first select a word from a seed pool as the forbidden word. Then, we create a question that ensures the respondent will inevitably talk about the forbidden word. At test time, a prompt with a \taskD combines an instruction forbidding discussion of a certain word and a question that prompts the LLM to engage with that word.

\subsubsection{Vision-Language (V-L) Conflict}
\label{subsec:VL}
refers to conflicts between the multimodal components of vision and language. Below will elaborate on 4 subclasses of conflicts: \taskE, \taskF, \taskG, and \taskH.

\paragraph{\taskEpara} consists of two conflicting instructions respectively in vision and language form, as presented in Fig.~\ref{fig:task5}. The generation of \taskE can be summarized in two steps. First, a list of short sentences is generated to provide the context for the conflicts. Second, utilizing instructions pairs from Section~\ref{par:exclusive}, an image of the concatenation of an instruction and a sentence is crafted. The image varies in font, size, and color to augment diversity.
At test time, presenting the image and the conflicting instruction concurrently yields a conflict.

\paragraph{\taskFpara} involves a simple chart with an incorrect text description, as shown in Fig.~\ref{fig:task6}. It is created through four steps. First, a list of commonly used words and entities with related numerical data is generated to decide the conflict's topic. Second, a narrative description and question are crafted for each entity and its data. Third, the numerical data is manipulated by changing the maximum value to the minimum value. Finally, a chart is plotted based on the altered data, with random choices for font, size, color, and other style options. At test time, combining the question and the figure creates a \taskF.

\paragraph{\taskGpara} involves an image of geometric shapes with an incorrect description, as shown in Fig.~\ref{fig:task7}. The generation process has four main steps. First, an image of two geometric objects with different attributes (shape, size, color, and position) is created. Second, a phrase is crafted to describe an object using two attributes (e.g., "the smaller gray object"). Third, this phrase is modified to refer to a non-existent object (e.g., "the larger gray object"). Finally, a question is generated about a third attribute of the non-existent object (e.g., "What is the shape of the larger gray object?"). At test time, presenting the image and question together creates a \taskG.

\paragraph{\taskHpara} involves an erroneously classified image, as shown in Fig.~\ref{fig:task8}. To be specific, a question about the wrong class (\eg, ``kiwi") should be answered according to the given image (\eg, ``ostrich"). The generation process of \taskH is based on the ImageNet dataset~\cite{deng2009imagenet}. First, we generate some questions about a label and retrieve images according to that label in the ImageNet dataset. Second, we substitute the correct label in the questions with some similar but different objects. At test time, combining the image and the substituted question will create a conflict. 

\begin{tcolorbox}[
    colback=yellow!10!white,
    colframe=blue!60!black,
    title=\taskH,
    center title,
    rounded corners,
    fonttitle=\bfseries,
    breakable
]
\textbf{Substitute object}: Ostrich to Kiwi

\begin{center}
    \includegraphics[width=0.5\linewidth]{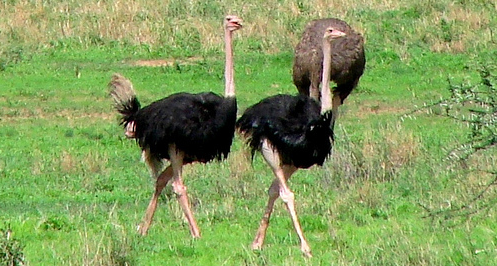}
\end{center}

\textbf{Question}: Does the picture depict the \textcolor{darkred}{kiwi's size}?
\end{tcolorbox}

\section{Approach}
\label{sec:approach}
In this section, we delve into our exploration using in-context learning techniques, detailed in Section~\ref{sec:ICL}.
Through experiments across various Large Multimodal Models (LMMs), we've pinpointed a crucial challenge where LMMs struggle to detect instruction conflicts.
Additionally, we introduce our proposed Cognitive Awakening Prompting (\ourApproach) approach, outlined in Section~\ref{sec:Cognitive Awakening Prompting}.

\subsection{In-Context Learning}
\label{sec:ICL}

We study three in-context learning techniques in \ourMethod, including few-shot prompting~\cite{brown2020language}, zero-shot chain-of-thoughts prompting~\cite{NEURIPS2022_8bb0d291}, and self-consistency prompting~\cite{wang2022self}.
Although few-shot prompting has been widely used in Large Language Models, its application in Large Multimodal Models (LMMs) remains limited.
Recent research highlights challenges such as LMMs' inability to support multiple image inputs or comprehend sophisticated few-shot prompts~\cite {zhao2023mmicl,jiao2024enhancing}.
Consequently, few-shot prompting is primarily employed within the language-language paradigm.
Here, we detail the application of these prompting techniques in our \ourMethod.

\emph{Zero-shot Prompting} refers to the ability of the model to perform a task without providing examples of how to perform a task correctly.
We task the model with generating responses in \ourMethod solely based on its general knowledge and understanding of language and vision.
This capability underscores the model's innate capacity to detect self-contradictory conflicts.

\emph{Zero-shot Chain-of-thoughts Prompting~\cite{wei2022chain} (CoT)} involves appending text like ``Please think step by step'' to user prompts, proven to enhance LMMs' inference ability. 
 In our experiment, we incorporate this text into the prompt.

\emph{Self-consistency Prompting~\cite{wang2022self} (SC)} involves sampling multiple reasoning paths and selecting the most consistent answers.
In this paper, we generate three replies for each instruction (3-SC) and determine the final result through majority voting.

\subsection{Cognitive Awakening Prompting}
\label{sec:Cognitive Awakening Prompting}

Our initial exploration reveals an intriguing phenomenon: the performance order in vision-language tasks is 0-Shot, CoT, and 3-SC across diverse LMMs, shown in the Appendix.
While 3-SC provides \emph{additional experience} through more attempts, CoT offers \emph{extra knowledge} by stimulating reasoning capabilities through a chain of thought. 
However, neither surpasses the simplicity of zero-shot prompting, suggesting that both \emph{additional experience} and \emph{extra knowledge} derived from the model itself may be counterproductive.
We hypothesize that the LMMs may not fully grasp \emph{restricted cognition} in the self-contradictory instruction scenarios.

Therefore, we propose a plug-and-play prompting approach to infuse cognition from the external world: Cognitive Awakening Prompting (\ourApproach).
The externally added cognition prompt reminds LMMs of potential inconsistencies hidden in their cognition, \eg, adding ``Please be careful as there may be inconsistency in user input. Feel free to point it out." at the end of the prompt.
The injected cognition does not impair the basic functioning of LMMs but fosters self-awareness of internal information and knowledge defects.
Detailed experiments are presented in Section~\ref{sec:vision-language-exp}.

While \ourApproach stems from observation and analysis in vision-language tasks, it also demonstrates promise in language-language tasks, outperforming 3-Shot in over half of LMMs.
Generally, the 3-Shot provides \emph{extra information} since more question-answer pairs are provided.
This underscores that \emph{cognition} represents a higher level of existence than \emph{experience}, \emph{information}, and \emph{knowledge}.
\ourApproach embodies a prompting technique standing on the cognition dimension, enabling the identification of LMMs' shortcomings and exploration of profound issues. 
Detailed experiments are outlined in Section~\ref{sec:language-language-exp}.

\section{Experiments}

In this section, we begin with the experimental settings and introduce the Large Multimodal Models (LMMs), metric, and evaluation in Section~\ref{sec:experimental settings}. 
Furthermore, we assess the capacity of various large multimodal models (LMMs) to detect self-contradictory instructions in \ourMethod for language-language (L-L) and vision-language (V-L) tasks, in Section~\ref{sec:language-language-exp} and Section~\ref{sec:vision-language-exp} respectively.

\subsection{Experimental Settings}
\label{sec:experimental settings}
\paragraph{Large Multimodal Models} including 11 types are experimented on \ourMethod to assess how well LMMs can detect self-contradictory instructions.
To elaborate, L-L conflicts are experimented on ChatGLM~\cite{chatglm}, ChatGPT~\cite{chatgpt}, GPT-4~\cite{openai2023gpt4}, Llama 2~\cite{llama2}, and GLM-4~\cite{glm4}.
V-L conflicts are experimented on GPT-4V~\cite{openai2023gpt4}, LLaVA-1.5~\cite{liu2023improvedllava}, Gemini~\cite{geminiteam2023gemini}, LLaMA-Adapter V2~\cite{gao2023llamaadaptorv2}, BLIP-2~\cite{li2023blip2}, and SPHINX-v2~\cite{lin2023sphinx}.

\begin{table}[htbp]
\centering
\renewcommand{\arraystretch}{1.2}
\caption{\textbf{Evaluation of LLM agents aligns with human experts.} Spearman correlation coefficient and Concordance rate are calculated between the evaluation results of the LLM agents and the human experts on vision-language conflicts.
}

\adjustbox{max width=0.95\linewidth}{
\begin{tabular}{ c  c  c }
    \toprule

    Reply LMM &   Spearman's $\rho$ & Concordance\\
    \midrule    

   GPT-4V &     0.881           & 94\%       \\ 
LLaVA-1.5 &     0.999           & 99\%       \\ 
   Gemini &     0.854           & 97\%      \\

     \bottomrule
\end{tabular}
}
\label{tab:main-vision-eval}
\end{table}

\paragraph{Metric} in our experiment is the hit ratio, which is defined as the proportion of the conflict-aware replies with the total replies.
To calculate the hit ratio, each reply generated by LMM will be evaluated to determine whether it successfully identifies the conflict hidden in the user's input.

\paragraph{Evaluation} is first conducted by human experts who can provide the most accurate evaluation. 
However, it is prohibitively costly to evaluate data manually in large-scale experiments.
Employing LLMs as an evaluation agent offers a more efficient and cost-effective alternative.
An experiment further demonstrates that LLMs as evaluation agents align with human experts, shown in Table~\ref{tab:main-vision-eval}.
In our experiment, a uniform prompt for all tasks is designed to prompt LLMs as evaluation agents. 
Initially, a set of replies generated by LMM on \ourMethodCore was collected. These replies were then evaluated by both human experts and GPT-4~\cite{openai2023gpt4}. Spearman correlation coefficient and concordance rate are calculated to measure the evaluation consistency between humans and LLM. As recorded in Table~\ref{tab:main-vision-eval}, GPT-4 demonstrates a close alignment to human evaluative standards.

\begin{table}[ht]
\centering
\renewcommand{\arraystretch}{1.2}
\caption{\textbf{Our \ourApproach significantly improves the performance of detecting instruction conflicts on \ourMethod.}
Scores in the table are hit ratios evaluated by ChatGPT. The higher, the better. * means tested on \ourMethodBase introduced in Section~\ref{subsec:ourmethod}.}

\adjustbox {max width=0.92\linewidth}{
\begin{tabular}{l c c c c c}
    \toprule
    Model& \taskA & \taskB & \taskC & \taskD & Total \\
    \midrule
    ChatGLM&            21.9\%&  9.0\%&  9.9\%& 27.6\%& 17.1\%\\
        \quad + CoT    &38.9\%& 11.4\%&  5.8\%& 42.6\%& 24.7\%\\
        \quad + 3-Shot &48.4\%&  9.1\%& \textbf{17.9\%} & \textbf{95.2\%}&  42.6\%\\
        \rowcolor{lightgray} \quad + \ourApproach & \textbf{69.1}\%& \textbf{70.2\%} & 17.0\%&48.1\%&\textbf{51.1\%}\\
    \midrule
    ChatGPT&            35.7\%& 13.4\%&  4.4\%&  1.8\%& 13.8\% \\
        \quad + CoT    &36.9\%& 25.0\%& 10.4\%&  2.1\%& 18.6\% \\
        \quad + 3-Shot &71.4\%& 22.4\%& \textbf{28.8\%}& \textbf{ 4.5\%}& 31.8\% \\
        \rowcolor{lightgray} \quad + \ourApproach &\textbf{80.9\%}&\textbf{66.6\%}&11.6\%&1.8\%&\textbf{40.2\%} \\
    \midrule
    GLM-4&              31.1\%& 33.0\%& 20.6\%& 52.0\%& 34.2\%\\
        \quad + CoT &   33.4\%& 49.3\%& 25.4\%& 53.0\%& 40.3\%\\
        \quad + 3-Shot &\textbf{50.8\%}& 45.9\%& 52.8\%& 67.3\%& 54.2\%\\
        \rowcolor{lightgray} \quad + \ourApproach & 49.8\%& \textbf{84.4\%}& \textbf{54.5\%}& \textbf{83.9\%}& \textbf{68.1\%}\\
    \midrule
    Llama2 &         46.6\%& 26.9\%&  8.4\%& 21.2\%& 25.8\%\\
        \quad + CoT &   44.8\%& 29.7\%&  8.0\%& 18.5\%& 25.2\%\\
        \quad + 3-Shot &17.8\%& \textbf{75.8\%}& \textbf{31.8\%}& \textbf{52.2\%}& \textbf{44.4\%}\\
        \rowcolor{lightgray} \quad + \ourApproach & \textbf{67.8\%}& 43.8\%& 6.7\%& 19.4\%& 34.4\%\\
    \midrule
    GPT-$4^*$ &         28.4\%& 26.8\%& 13.2\%& 42.0\%& 27.6\%\\
        \quad + CoT &   25.6\%& 40.0\%& 29.2\%& 90.4\%& 46.3\%\\
        \quad + 3-Shot& \textbf{90.0\%}& 68.0\%& \textbf{70.8\%}& \textbf{98.4\%}& \textbf{81.8}\%\\
        \rowcolor{lightgray} \quad + \ourApproach & 74.4\%& \textbf{96.0\%}& 26.0\%& 91.6\%& 72.0\%\\
    \bottomrule
\end{tabular}
}

\label{tab:language-language-tasks}
\end{table}

\subsection{Language-Language Conflict}
\label{sec:language-language-exp}
We experiment with ChatGPT, ChatGLM, GLM-4, and Llama2-7b-chat on the \ourMethod, while GPT-4 is tested on \ourMethodBase, as introduced in Section~\ref{subsec:ourmethod}. For prompt setting, we apply zero-shot, chain-of-thoughts, few-shot, and cognitive awakening prompting in experiments on language-language conflict.

Table \ref{tab:language-language-tasks} demonstrates the performance of an LMM under different prompt settings. Existing LMMs perform poorly in handling language-language conflicts. However, in-context learning techniques can improve the performance of LMMs to a different extent. Chain-of-thoughts prompting offers a relatively modest increase in the hit ratio, approximately by a factor of 1.5. This relatively moderate improvement of chain-of-thoughts prompting may result from the fact that it can be counted as part of the conflicting instructions, thus failing to fully elicit the reasoning ability of LMMs. Few-shot and our \ourApproach prompting can significantly improve LMM's overall hit ratio, approximately doubling or tripling their performance. This may be due to the external message that reminds LMMs of the potential existence of conflicts. In practice, few-shot and \ourApproach can be combined to further improve LMMs' awareness of dissonance.

It's also noteworthy that different tasks vary in difficulty. Specifically, \taskC seems to be more challenging to most LMMs, showing LMMs' inability to understand the exclusion of two instructions. \taskA and \taskB are relatively easier for LMMs and that may result from the powerful information retrieval ability of LMMs. LMMs' performances vary on \taskD, while GPT-4 achieves a hit ratio of 98.4\%, ChatGPT only achieves 4.5\%. This might be due to the difference in their training data.

\begin{table}[ht]
\centering
\renewcommand{\arraystretch}{1.2}
\caption{\textbf{GPT-4V outperforms other LMMs greatly in all tasks on \ourMethodCore.} LLaMA-A2 represents the LLaMA-Adapter V2. The replies are evaluated by human experts for more precise results.}

\adjustbox{max width=0.9\linewidth}{
\begin{tabular}{c c c c c a}
    \toprule
    Model  &\taskE & \taskF & \taskG & \taskH & \textbf{Total}\\
    \midrule
    BLIP-2 &   0.0\% &   0.0\% &    0.0\%&    0.0\%& 0.0\%  \\
    LLaMA-A2 &0.0\%&  0.0\% &    0.0\%&    0.0\%& 0.0\%  \\
    LLaVA-1.5 &   0.0\% &   0.0\% &    0.0\%&    0.0\%& 0.0\%  \\ 
    SPHINX-v2 &   0.0\% &   0.0\% &    0.0\%&    2.0\%& 1.0\% \\
    Gemini   &   6.7\% &   0.0\% &    0.0\%&   20.0\%& 11.0\% \\ 
    GPT-4V &  \textbf{80.0\%} &  \textbf{33.3\%} &   \textbf{40.0\%}&   \textbf{68.0\%}& \textbf{59.0\%} \\
    \bottomrule
\end{tabular}
}

\label{tab:main-vision-reply}
\end{table}

\subsection{Vision-Language Conflict}
\label{sec:vision-language-exp}
We experiment with GPT-4V, LLaVA-1.5 (with 8-bit approximation), and Gemini~\footnote{The experiments utilize the website version of Gemini.}, LLaMA-Adapter V2 (BIAS-7B), BLIP-2 (FlanT5$_{XXL}$), and SPHINX-v2 on \ourMethodCore using basic zero-shot prompting. As evident from Table~\ref{tab:main-vision-reply}, GPT-4V outperforms other LMMs greatly in all 4 tasks. Even SPHINX performs miserably, a rather large open-source model. Gemini shows a slightly better result but the overall performance is still poor. This proves current LMMs' inability to detect self-contradictory instructions. Considering the unparalleled advantage of GPT-4V, we reckon the simple design of current open-source LMMs cannot handle self-contradictory instructions correctly even with LLMs, and more advanced architecture is a must to handle such a challenge.

It is also noteworthy that \taskE and \taskH are relatively easy for GPT-4V and Gemini to perform, while \taskF and \taskG exhibit the greatest difficulty. This demonstrates that current LLMs still struggle with interpreting figures and performing spatial reasoning tasks.

We further the experiment to explore whether in-context learning can improve performance. Due to the current limitations of vision-language models, it is typically not recommended to apply few-shot learning in this setting as we've discussed in Section~\ref{sec:ICL}. We simply apply plain zero-shot prompting, zero-shot chain-of-thoughts prompting~\cite{NEURIPS2022_8bb0d291}, self-consistency prompting~\cite{wang2022self}, and cognitive awakening prompting.

\begin{figure}[htbp]
    \centering
    \includegraphics[width=0.8\linewidth]{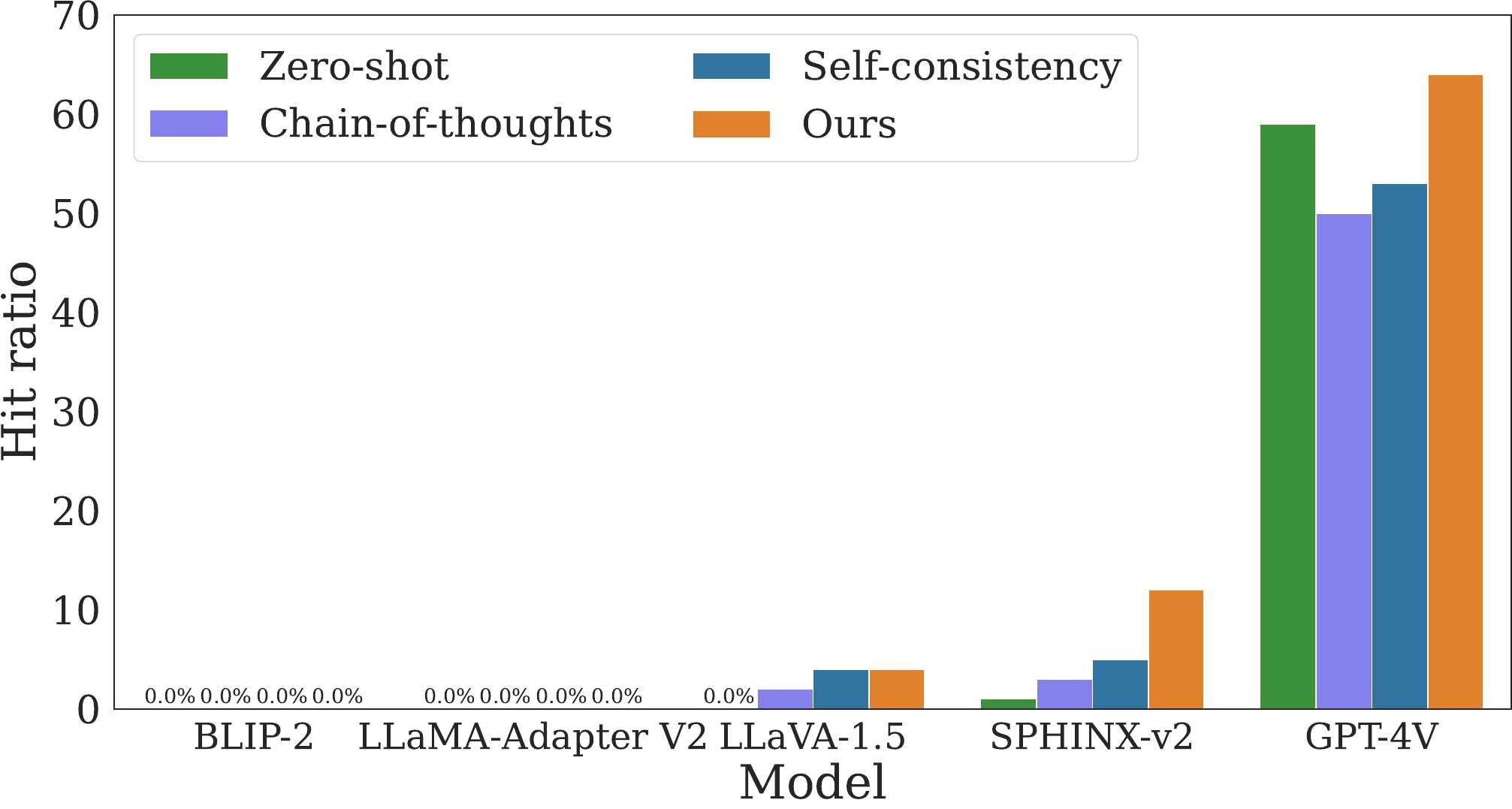}
    \caption{\textbf{\ourApproach improves LMMs' performance greatly on \ourMethodCore.} Chain-of-thoughts and self-consistency prompting bring limited improvement. Replies are evaluated by human experts for more precise results.
}
    
    \label{fig:method_vl}
\end{figure}

Fig.~\ref{fig:method_vl} shows that \ourApproach greatly enhances LMMs' performance. This is most evident in SPHINX-v2, where \ourApproach raises its hit ratio from a poor 1.0\% to a commendable 12.0\%. This improvement applies to LLaVA-1.5 and GPT-4V where \ourApproach constantly outperforms in-context learning skills like chain-of-thoughts and self-consistency, showing the indisputable superiority of \ourApproach. BLIP-2 and LLaMA-Adapter V2 cannot detect any self-contradictory instruction no matter the in-context learning skills we apply, and we reckon that the base LLMs they use may not be powerful enough to handle such a challenging problem (FlanT5 and LLaMA-7b respectively). Unlike in the language-language setting, chain-of-thoughts prompting only brings limited improvement on LLaVA-1.5 and even a negative effect on GPT-4V's performance. Self-consistency prompting, as an improved version of chain-of-thoughts prompting, shows a similar but slightly more satisfying result than CoT prompting.

It is also worth mentioning that, in our self-contradictory setting, these in-context learning skills sometimes fail to achieve the originally expected result, which could be the reason why they fail to improve performance on \ourMethod. For example, ``Please think step by step'' is meant to elicit a chain of LMM thoughts but is sometimes deemed as a normal context to be translated, paraphrased, and summarized in \taskE. 

Finally, \ourApproach is \emph{harmless} since it serves as an additional module for conflict detection.
If it detects conflicts, it can ask the user to check input.
Otherwise, the original task will proceed as usual.
We conduct experiments on two non-conflict datasets to prove that \ourMethod will not lead to misjudgment in normal cases, MMMU~\cite{yue2024mmmu} by LLaMa-Adapter-V2~\cite{gao2023llamaadaptorv2} and MMLU~\cite{hendrycks2020measuring} by GPT-4~\cite{openai2023gpt4}.
We find that only 1.38\% replies mistakenly mentioned a conflict on the MMMU benchmark (1.11\% on the MMLU).

\section{Conclusion}
\label{sec:conclusion}

We introduce the Self-Contradictory Instructions (\ourMethod) benchmark, comprising 20,000 conflicts distributed between language and vision domains.
This benchmark aims to evaluate Large Multimodal Models (LMMs) regarding their ability to detect conflicting commands.
Our innovative automatic dataset creation framework, \ourTool, facilitates this process and encompasses a wide range of instruction complexities.
Our evaluation reveals current LMMs' consistent struggle to identify instruction conflicts.
Hence, we propose a novel approach, Cognitive Awakening Prompting (\ourApproach), to inject cognition from the external world, leading to a substantial improvement in dissonance detection.

\paragraph{Social Impact} 

Our work on the \ourMethod benchmark, along with the \ourTool framework and \ourApproach approach, has significant social implications. It provides researchers and practitioners with a standardized platform to assess and enhance LMMs' ability to navigate conflicting instructions, advancing human-computer interaction and communication technologies. The \ourTool framework facilitates the creation of diverse instruction datasets, promoting inclusivity in AI research. Additionally, the \ourApproach approach integrates external cognition into multimodal models, enhancing context-aware understanding. By improving dissonance detection, our approach boosts LMM performance and fosters trust and reliability in AI systems, essential for societal integration.

\paragraph{Limitations} To begin with, we only include language-language and vision-language paradigms. More modalities will be included in our \ourMethod benchmark based on our automatic framework, \ourTool. 
Besides, no fine-grained control is introduced to determine the conflict degree.
Finally, we do not provide a detailed study on the attention mechanism of large multimodal models when confronted with conflict instructions.

\section*{Acknowledgements}

This research is supported by the Key R\&D Program of Shandong Province, China (2023CXGC010112). We express our gratitude to the funding agency for their support.

\bibliographystyle{splncs04}
\bibliography{main}

\newpage
\appendix
\setcounter{page}{1}

\section{\ourMethod Construction}
\label{sec:appendix_construction}

This section provides a detailed description of \ourMethod automatic generation, with individual explanations for the eight tasks in \ourMethod.

\subsection{Language-Language Conflict}

Language-language conflicts are categorized into four distinct tasks: \taskA, \taskB, \taskC, and \taskD. Their generation processes will be detailed separately in the following sections.

\subsubsection{\taskA}
\taskA generation involves a systematic process that can be divided into three key steps:

\begin{enumerate}%
    \item \textbf{Develop a Context}: Start by crafting a contextual setting that establishes a strict rule and provides background information. This context serves as the foundation for the subsequent conflict generation.
    
    \item \textbf{Generate a Violating Sentence}: Create a sentence that intentionally violates the established rule, as if it is acceptable to break the rule within the given context. This violating sentence should effectively challenge the rule's integrity.
    
    \item \textbf{Pose an Unanswerable Question}: Formulate a single question that becomes unanswerable when posed to the model due to the paradox created by the rule violation. The question should be designed to make it impossible for the model to provide a coherent or correct response while confronting the conflict introduced by the rule violation.
\end{enumerate}

\newpage 

\begin{tcolorbox}[
    colback=yellow!10!white,
    colframe=red!60!black,
    title=\taskA,
    center title,
    rounded corners,
    fonttitle=\bfseries,
    breakable
]
\textbf{Rule}: \textcolor{darkgreen}{City A has only 1 mayor, Megan, from 2012 to 2020.}

\textbf{Violation}:
\textcolor{darkred}{Leon gave a talk in 2015 as the mayor of City A.}

\textbf{Question}: Who served as the mayor of City A in 2015?
\end{tcolorbox}

\subsubsection{\taskB}
\taskB introduces a distinct type of L-L conflict related to the attributes of fictitious objects. Its generation comprises three key steps:

\begin{enumerate}%
    \item \textbf{Generate Object Description}: Prompt LLM to create a descriptive text about a fictitious object, including various attributes that the virtual object supposedly possesses. This text should describe the object in detail, even though it does not exist in the real world.
    
    \item \textbf{Attribute Description Extraction}: Prompt LLM to extract descriptions for each attribute mentioned in the generated text. Each attribute description will be used to generate a conflict in the next step.
    
    \item \textbf{Generate Contradictory Sentence}: For each attribute extracted in the previous step, we instruct LLM to generate a single sentence that conveys the exact opposite of the original attribute description. This sentence will contradict the original description.
\end{enumerate}

Following the procedures outlined above, we generate a virtual object with a descriptive text that enumerates its various attributes. Each attribute has a contradictory sentence that disputes the original text. By simply concatenating each contradictory sentence with the original text, we can generate an object attribute conflict.

\begin{tcolorbox}[
    colback=yellow!10!white,
    colframe=red!60!black,
    title=\taskB,
    center title,
    rounded corners,
    fonttitle=\bfseries,
    breakable
]
\label{box:task2}
\textbf{Descriptive text}: The Etherian LuminoSphere is a wondrous object that exists beyond reality. Its form is that … \textcolor{darkred}{radiating an ever-changing spectrum of colors} …

\textbf{Attribute}: \textcolor{darkred}{color}

\textbf{Original description}: \textcolor{darkred}{It radiates an everchanging spectrum of colors}

\textbf{Opposite description}: \textcolor{darkgreen}{It emits a constant single color.}

\textbf{Instruction}: Describe the \textcolor{darkred}{color} of the Etherian LuminoSphere.
\end{tcolorbox}

\subsubsection{\taskC}
In the case of \taskC, users provide two conflicting tasks within the same prompt. \label{par:appendix_exclusive}
The generation process is detailed below.
\begin{enumerate}%
    \item \textbf{Generate Exclusive Instruction pairs }: Handcraft a list of exclusive instructions and prompt LLM to generate more (\eg,``Translate the text to Chinese" versus ``Translate the text to French"). Then, choose desired pairs and add them back to the list and thus iteratively generate more. The list can be sampled to make conflicts.
    
    \item \textbf{Generate Task Text}: Start with prompting LLM to generate a pool of common elements in a story. Then randomly sample 3 elements from the pool and instruct LLM to create a short storyline consisting of these elements.
\end{enumerate}

By combining the exclusive instructions and the text, the final conflict is generated. For example, a prompt may be formatted as follows:
$$\left\{ \{instruction1\}\{text\}\{instruction2\} \right\}$$

\begin{tcolorbox}[
    colback=yellow!10!white,
    colframe=red!60!black,
    title=\taskC,
    center title,
    rounded corners,
    fonttitle=\bfseries,
    breakable
]
\label{box:task3}
\textbf{Instruction 1}: \textcolor{darkgreen}{Please translate anything below to Chinese.}

\textbf{Paragraph}: Alice was beginning to get tired of sitting by her sister on the bank, and of having nothing to do: once or twice she had peeped into…

\textbf{Instruction 2}: \textcolor{darkred}{Please translate anything above to French.}
\end{tcolorbox}

\subsubsection{\taskD}
\taskD initially provide LLM with an instruction to avoid using a specific word, referred to as the ``forbidden word" (\eg, word A). Subsequently, users ask LLM a question or request information that inherently requires the use of the forbidden word to answer accurately. We apply the following approach to synthesize the conflict.

\begin{enumerate}%
    \item \textbf{Sample from categories}: First, instruct LLM to generate a list of different categories(\eg, "history", "chemistry"). For each different category, let LLM sample 50 or more different entities.
    
    \item \textbf{Pose Unique Question}: For each entity sampled, we prompt LLM to generate a question that can only be correctly answered by the entity.

    \item \textbf{Select the questions}: Use an LLM agent to judge whether the answers to these generated questions are unique. This is necessary because LLM tends to return questions with ambiguous answers even with explicit emphasis on uniqueness.
\end{enumerate}

To create conflicts, simply pose one question while asking LLM not to mention the entity corresponding to it.

\begin{tcolorbox}[
    colback=yellow!10!white,
    colframe=red!60!black,
    title=\taskD,
    center title,
    rounded corners,
    fonttitle=\bfseries,
    breakable
]
\label{box:task4}
\textbf{Forbidden word}: \textcolor{darkgreen}{Cuba}

\textbf{Restriction}: Directly answer the following question. \textcolor{darkred}{You MUST NOT mention Cuba.}

\textbf{Question}: What is the \textcolor{darkred}{Caribbean island nation that has a communist government?}
\end{tcolorbox}

\subsection{Vision-Language Conflict}
Vision-Lanuage conflict refers to conflicts between multimodal. Below will elaborate on 4 subclasses of conflicts: \taskE, \taskF, \taskG, and \taskH.

\subsubsection{\taskE} 
\taskE mostly inherits from the \taskC in L-L conflict but presents in a language-vision form. The generation process can be summarized in 2 steps below.
\begin{enumerate}%
    \item \textbf{Generate short sentence}: Instruct LLM to generate a list of meaningful sentences instead of long stories.
    
    \item \textbf{Create Image}: Directly use instructions pair from \taskC, concatenating one of the instructions with the sentence. Generate an image of solely the text, varying in font, size, and color.

\end{enumerate}
To generate a conflict, simply simultaneously give an image and a conflicting instruction of the image to LMM.

\begin{tcolorbox}[
    colback=yellow!10!white,
    colframe=blue!60!black,
    title=\taskE,
    center title,
    rounded corners,
    fonttitle=\bfseries,
    breakable
]
\label{box:task5}
\textbf{Instruction 1}: 

\vspace{0.2cm}
\includegraphics[width=\linewidth]{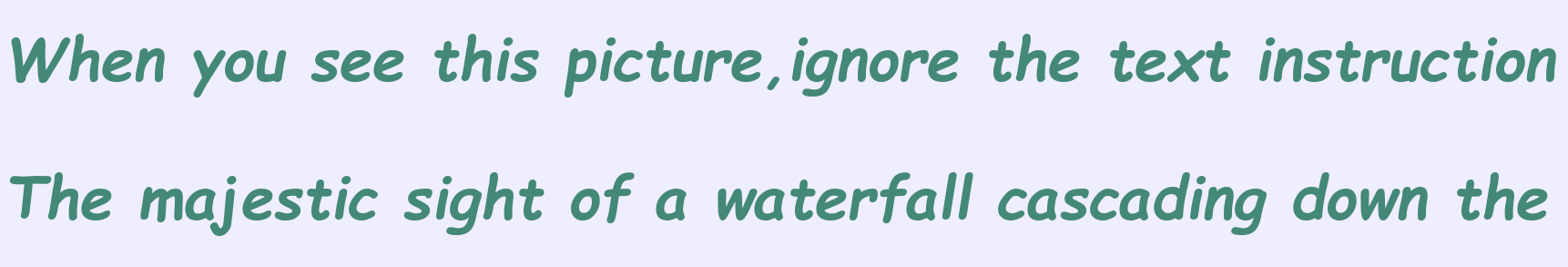}

\textbf{Instruction 2}: \textcolor{darkred}{Paraphrase the given sentence in a poetic way.}
\end{tcolorbox}

\subsubsection{\taskF}
\taskF examines LMM's ability to read figures and relate them to text information. To elaborate, users provide LMM with both a figure and some text description contrasting with it when asking a related question. The generation process is detailed below.
\begin{enumerate}%
    \item \textbf{Generate Data Dictionary}: First get a list of 500 commonly used words and entities in English. For each word, instruct LLM to return a JSON format of a dictionary that can be used to plot a bar graph, pie graph, or line chart.
    
    \item \textbf{Describe Data}: Feed the data dictionary to an LLM and ask it to describe the data and make some conclusions. Then, prompt LLM to ask a question regarding the largest value in the dictionary.

    \item \textbf{Modify and Plot}: Tamper with the data dictionary, \ie change the largest value in it to the smallest value. Plot corresponding figures based on the tampered data, randomly choosing font, size, color and other stylish options.
\end{enumerate}

Simply concatenating the question and the figure yields a conflict.

\begin{tcolorbox}[
    colback=yellow!10!white,
    colframe=blue!60!black,
    title=\taskF,
    center title,
    rounded corners,
    fonttitle=\bfseries,
    breakable
]
\label{box:task6}
\textbf{Type}: Line chart

\vspace{0.2cm}
\includegraphics[width=\linewidth]{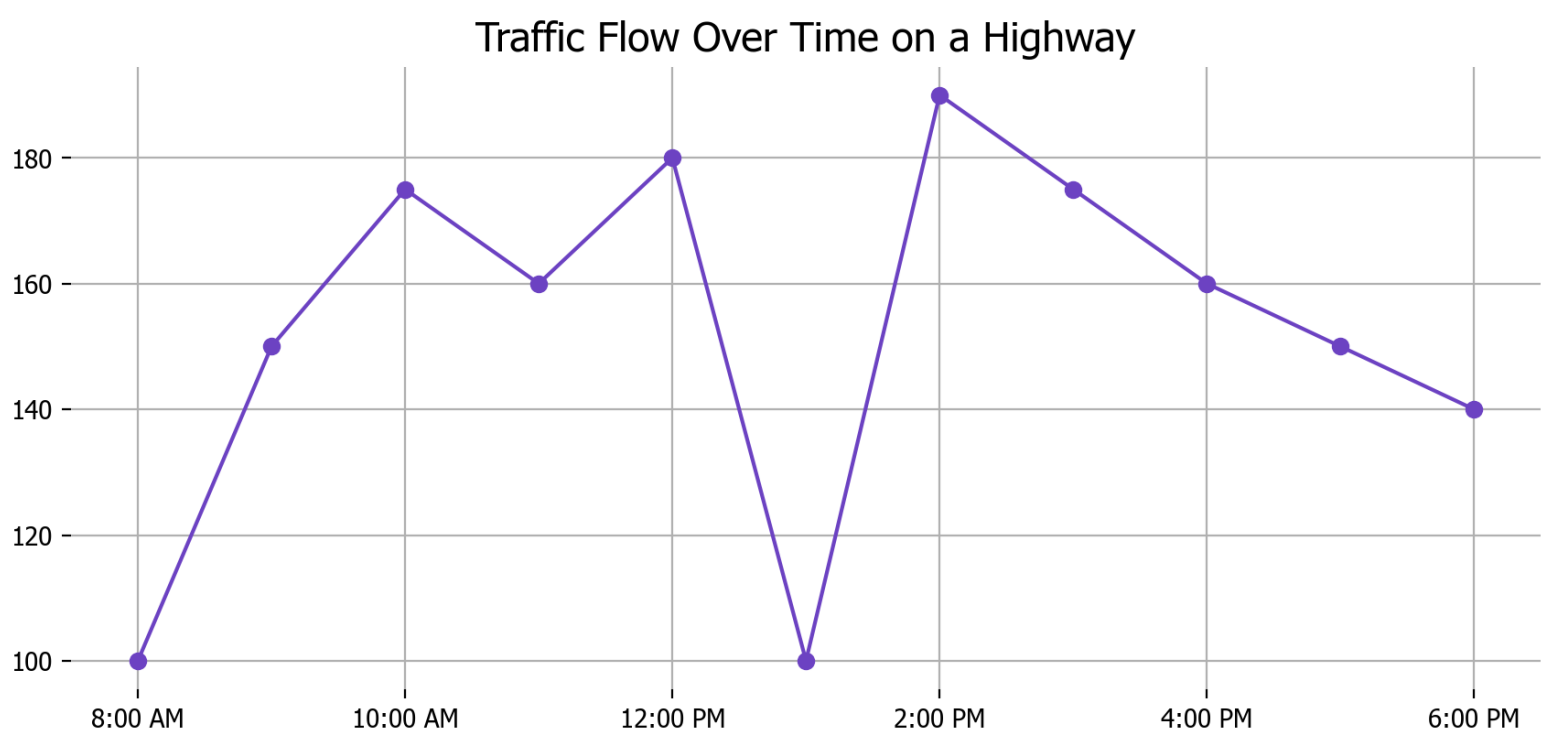}

\textbf{Description}: On a typical day, the traffic flow on a highway starts at 8:00 AM with 100 vehicles and gradually increases, \textcolor{darkred}{its peak at 1:00 PM} with 200 vehicles. After that, the traffic begins to subside...

\textbf{Question}: At what time did the traffic flow \textcolor{darkred}{reach its peak on the highway} based on the given data?
\end{tcolorbox}

\subsubsection{\taskG}
\taskG challenges LMM's ability to detect dissonance between geometric objects and related text descriptions. In this setting, LMM is given an image of 2 geometric objects with certain colors and shapes. The generation process contains the following steps.
\begin{enumerate}%
    \item \textbf{Generate Shape}: Draw 2 random geometric objects, each with four attributes—shape, size, color, and position. 
    
    \item \textbf{Construct Question}: Query about one attribute while introducing confusion in two of the remaining attributes, \ie exchanging the description of 2 attributes in the text description.
\end{enumerate}

Conflict can be introduced by simply giving the image and the question to LMM. 

\newpage

\begin{tcolorbox}[
    colback=yellow!10!white,
    colframe=blue!60!black,
    title=\taskG,
    center title,
    rounded corners,
    fonttitle=\bfseries,
    breakable
]
\label{box:task7}
\textbf{Conficted attribute}: Color \& Size

\vspace{0.2cm}

\begin{center}
   \includegraphics[width=0.5\linewidth]{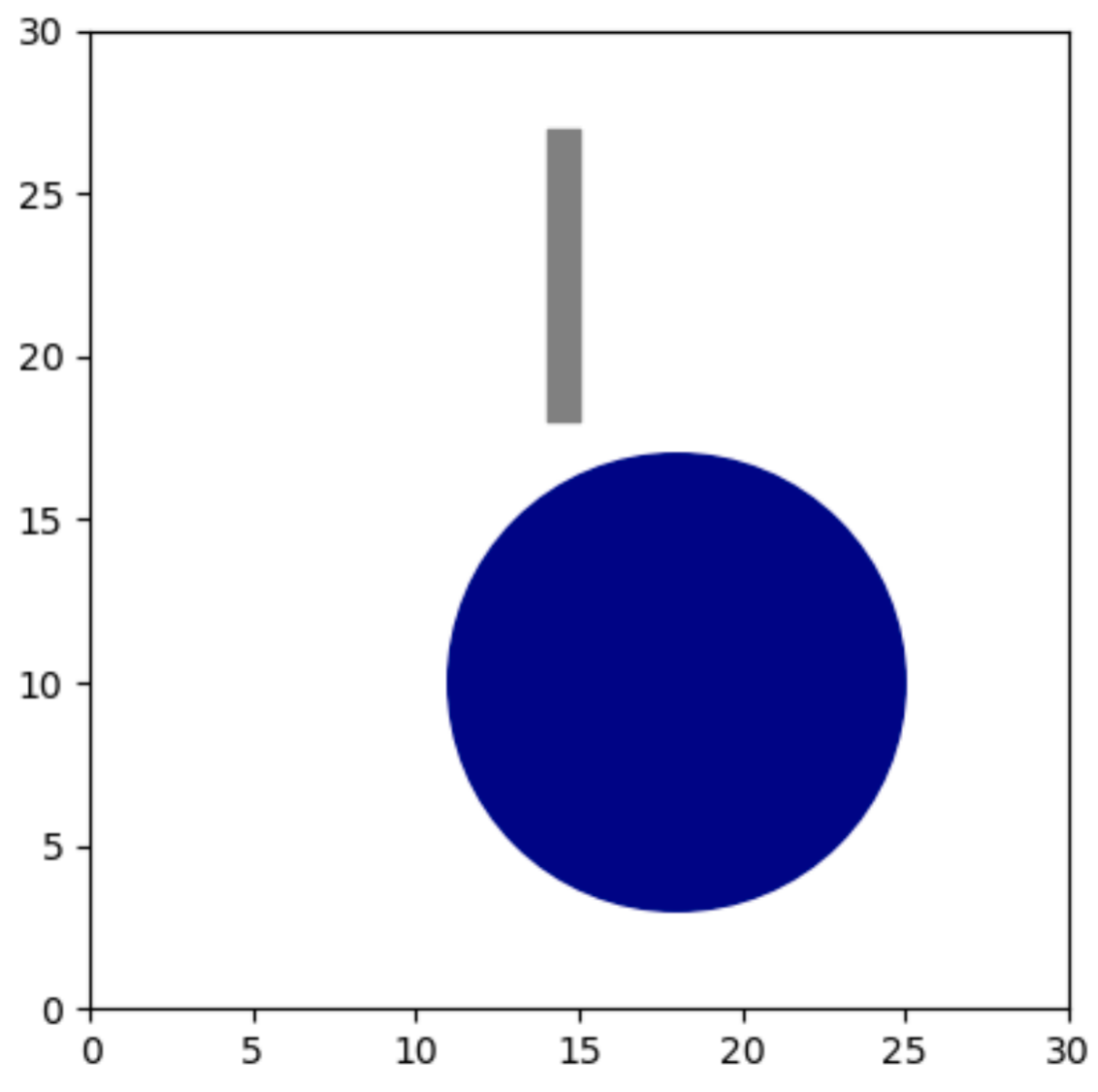} 
\end{center}

\textbf{Question}: What's the shape of the \textcolor{darkred}{larger gray} object?
\end{tcolorbox}

\subsubsection{\taskH}
\taskH refers to situations where the text description classifies the input image into the wrong class. To be specific, users may ask for information about object A(e.g. an ostrich) in an image of object B (e.g. a rooster). The ImageNet dataset is used to create the conflict. The detailed process is below:
\begin{enumerate}%
    \item \textbf{Generate Similar Object}: For each imagenet-1k class name(e.g. "mop"), prompt LLM to generate several similar objects(e.g. "duster"). These similar objects will later be used to substitute the class name.
    
    \item \textbf{Pose Related Question}: For each class name, prompt LLM to ask several potential questions as if an image of that class is given. The questions must contain the class name.

    \item \textbf{Substitue Object}: Substitute the class name in the questions with a random similar object.
    \item \textbf{Sample Image}: Use the class name to retrieve an image from imagenet-1k validation set.
\end{enumerate}

The final conflict is generated by combining the sampled image and the substituted question. 

\newpage
\begin{tcolorbox}[
    colback=yellow!10!white,
    colframe=blue!60!black,
    title=\taskH,
    center title,
    rounded corners,
    fonttitle=\bfseries,
    breakable
]
\textbf{Substitute object}: Ostrich to Kiwi

\begin{center}
    \includegraphics[width=0.5\linewidth]{fig/img/task8.png}
\end{center}

\textbf{Question}: Does the picture depict the \textcolor{darkred}{kiwi's size}?
\end{tcolorbox}

\subsection{Dataset Overview}

The SCI framework comprises 20,000 conflicts, evenly split between language-language conflicts and vision-language conflicts, each comprising 4 subsets. The dataset is split into 3 different levels: \ourMethodCore, \ourMethodBase, and \ourMethodAll to cater to different needs. 

Regarding the splitting of the dataset, subsets are selected manually with a size of 10\% (1\%) of \ourMethodAll. Notably, \taskC, \taskD, and \taskE require extra effort to guarantee diversity when used in \ourMethodCore.

\section{Experiment details}
\label{sec:appendix_experiment}

\subsubsection{V-L tasks}

\begin{table}[htbp]
\centering
\renewcommand{\arraystretch}{1.2}
\caption{\textbf{The trend of hit rates remains consistent across datasets of different scales.} The replies are tested on \textbf{LLaVA-1.5} and evaluated by \textbf{ChatGPT}. CoT is the chain of thought. 3-SC represents using 3 examples for voting in self-consistency. \ourApproach is our method,  Cognitive Awakening Prompting.}

\adjustbox {max width=\linewidth}{
\begin{tabular}{l c c c c c}
    \toprule
    Scale& \taskE & \taskF & \taskG & \taskH & Total \\
    \midrule
    \ourMethodCore &            \textbf{20.0}\%&  0.0\%&  0.0\%& 2.0\%& 4.0\%\\
        \quad + CoT    &0.0\%& \textbf{6.7\%}&  0.0\%& 2.0\%& 2.0\%\\
        \quad + 3-SC &0.0\%& \textbf{ 6.7\%}& 0.0\% & 0.0\%&  1.0\%\\
        \rowcolor{lightgray} \quad + \ourApproach & 0.0\%& 0.0\% & 0.0\%&\textbf{10.0\%}&\textbf{5.0\%} \\
    \midrule
    \ourMethodBase &            \textbf{8.0}\%&  0.7\%&  0.5\%& 2.4\%& 2.6\%\\
        \quad + CoT    &2.0\%&\textbf{ 4.7\%}&  \textbf{2.0}\%& 1.6\%& 2.2\%\\
        \quad + 3-SC &0.0\%&  1.3\%& 0.0\% & 0.8\%&  0.6\%\\
        \rowcolor{lightgray} \quad + \ourApproach & 5.3\%& 1.3\% & 1.5\%&\textbf{3.8\%}&\textbf{3.2\%} \\
    \midrule
    \ourMethodAll &            6.4\%&  3.0\%&  0.9\%& 3.2\%& 3.2\%\\
        \quad + CoT    &2.0\%& \textbf{4.0\%}&  \textbf{1.5\%}& 3.1\%& 2.7\%\\
        \quad + 3-SC &0.6\%&  0.9\%& 0.1\% & 1.7\%&  1.1\% \\
        \rowcolor{lightgray} \quad + \ourApproach & \textbf{7.9\%}& 3.8\% & 0.9\%&\textbf{4.8\%}&\textbf{4.4\%} \\
    \bottomrule
\end{tabular}
}

\label{tab:sup-vision-language-llava}
\end{table}

Besides experiments on \ourMethodCore in the main text, massive experiments are also conducted on \ourMethodCore, \ourMethodBase, and \ourMethodAll with LLaVA-1.5. Table~\ref{tab:sup-vision-language-llava} detailed the results. Despite some numerical disparities, the consistent trend persists across \ourMethodCore, \ourMethodBase, and \ourMethodAll. Specifically, chain-of-thoughts prompting enhances performance in \taskF and \taskG, potentially impacting negatively on \taskE and \taskH. \ourApproach significantly improves \taskH while also greatly enhancing performance across other tasks.

\subsection{Evaluation by different Agents}
For the evaluation of replies from LMMs, human evaluation is the most accurate and expensive approach for evaluation, while LLM evaluation is a less accurate but efficient approach. The main text has presented correlation coefficients from GPT-4, while this section will demonstrate more detailed results from more evaluation agents. For L-L tasks, experiments are conducted on \ourMethodAll and \ourMethodBase, and evaluation is conducted by ChatGPT and ChatGLM. For V-L tasks, experiments conducted on \ourMethodCore are evaluated by both human experts and three LLMs. 

\subsubsection{L-L tasks}
\begin{table}[htbp]
\centering
\renewcommand{\arraystretch}{1.2}
\caption{\textbf{Evaluation results by ChatGLM are close to those by ChatGPT.} Scores in the table are hit ratios evaluated by \textbf{ChatGLM}. The higher, the better. * means tested on \ourMethodBase.}

\vspace{-2mm}

\adjustbox {max width=\linewidth}{
\begin{tabular}{l c c c c c}
    \toprule
    Model& \taskA & \taskB & \taskC & \taskD & Total \\
    \midrule
    ChatGLM&            15.6\%&  3.5\%&  1.8\%& 22.0\%& 10.7\%\\
        \quad + CoT    &25.8\%& 17.9\%&  5.7\%& 33.3\%& 20.7\%\\
        \quad + 3-Shot &31.3\%&  6.1\%& 7.5\% & \textbf{85.9\%} &  32.7\%\\
        \rowcolor{lightgray} \quad + \ourApproach & \textbf{58.0}\%& \textbf{64.5\%} & \textbf{17.8\%} &42.5\%&\textbf{45.7\%}\\
    \midrule
    ChatGPT&            14.9\%& 7.3\%&  3.0\%&  0.6\%& 6.5\% \\
        \quad + CoT    &23.4\%& 13.9\%& 9.6\%&  0.8\%& 11.9\% \\
        \quad + 3-Shot &56.8\%& 18.0\%& \textbf{19.6\%}& \textbf{ 3.4\%}& 24.5\% \\
        \rowcolor{lightgray} \quad + \ourApproach &\textbf{74.1\%}&\textbf{62.2\%}&7.2\%&0.8\%&\textbf{36.1\%} \\
    \midrule
    GPT-$4^*$ &         16.0\%& 19.2\%& 11.2\%& 32.8\%& 19.8\%\\
        \quad + CoT &   25.2\%& 29.6\%& 31.6\%& 76.4\%& 40.7\%\\
        \quad + 3-Shot& \textbf{80.4\%}& 57.6\%& \textbf{72.8\%}& \textbf{98.0\%}& \textbf{77.2\%}\\
        \rowcolor{lightgray} \quad + \ourApproach & 71.2\%& \textbf{92.8\%}& 30.8\%& 79.2\%& 68.5\%\\
    \bottomrule
\end{tabular}
}

\label{tab:sup-language-language-glm-eval}
\end{table}

All results of L-L tasks in the main text are from the evaluation by ChatGPT. Table~\ref{tab:sup-language-language-glm-eval} presents the results from ChatGLM evaluation. Although the evaluation results of ChatGLM exhibit a slight numerical discrepancy compared to ChatGPT, they demonstrate a similar trend when compared across different methods.

\subsubsection{V-L tasks}
Human experts conduct all the evaluations of V-L tasks in the main text. Below will elaborate on evaluations conducted by GPT-4.
\begin{table}[htbp]
\centering
\renewcommand{\arraystretch}{1.2}
\caption{\textbf{Evaluation results by GPT-4 constantly align with human evaluation results.} 
Scores in the table are hit ratios tested on \textbf{GPT-4V} and evaluated by \textbf{human experts} and \textbf{GPT-4}. The higher, the better.}

\vspace{-2mm}

\adjustbox {max width=\linewidth}{
\begin{tabular}{l l c c c c c}
    \toprule
    Model& Evaluator &\taskE & \taskF & \taskG & \taskH & Total \\
    \midrule

    \multirow{2}{*}{GPT-4V} & Human &            80.0\%&  33.3\%&  40.0\%& 68.0\%& 59.0\%\\
     & GPT-4 &            80.0\%&  33.3\%&  35.0\%& 62.0\%& 55.0\%\\
     \midrule
     
    \multirow{2}{*}{GPT-4V + CoT}  & Human &    60.0\%&  33.3\%&  25.0\%& 62.0\%& 50.0\%\\
     & GPT-4&            66.7\%&  33.3\%&  30.0\%& 52.0\%& 47.0\%\\

     \midrule

     \multirow{2}{*}{GPT-4V + SC}  & Human &       80.0\%&  13.3\%&  30.0\%& 66.0\%& 53.0\%\\
     & GPT-4&            86.7\%&  13.3\%&  20.0\%& 58.0\%& 48.0\%\\
        
    \midrule
    
     \multirow{2}{*}{GPT-4V + \ourApproach}  & Human  &        80.0\%&  73.3\%&  30.0\%& 70.0\%& 64.0\%\\
     & GPT-4&            73.3\%&  73.3\%&  20.0\%& 72.0\%& 62.0\%\\
    \bottomrule
\end{tabular}
}

\label{tab:sup-vison-language-eval}
\end{table}

As can be seen in Table~\ref{tab:sup-vison-language-eval}, GPT-4 yields almost identical outcomes (overall discrepancy within 5\%) to those of human experts, underscoring its unparalleled capability to consistently perform such evaluations.

\section{LMM Responses}
\label{sec:appendix_examples}
This section showcases some examples illustrating how LMMs respond to \ourMethod tasks. Each example box contains a user prompt and several replies by various LMMs. In the user prompt, parts referencing a conflict are highlighted in brown font. In LMM replies, sentences that acknowledge the presence of a conflict are in dark green font, and sentences that neglect the conflict are in dark red font. 

\subsection{L-L tasks}
Four example boxes are demonstrated for \taskA, \taskB, \taskC, and \taskD respectively.
\begin{tcolorbox}[
    colback=yellow!10!white,
    colframe=red!60!black,
    title=\taskA,
    center title,
    rounded corners,
    fonttitle=\bfseries,
    breakable
]
\label{box:task1_reply}
\textbf{User}: 

\textcolor{black}{City A has only one mayor, and Megan holds the position from 2012 to 2020. Leon delivered a speech on Christmas Day in 2015 as the mayor of City A. Who served as the mayor of City A in 2015?}

\textbf{ChatGLM}: 

\textcolor{darkred}{Based on the information provided, Megan served as the mayor of City A from 2012 to 2020. Therefore, she was the mayor in 2015 as well.}

\textbf{ChatGPT}: 

\textcolor{darkred}{Megan served as the mayor of City A from 2012 to 2020. Therefore, Megan would have been the mayor of City A in 2015 when Leon delivered the Christmas Day speech.}

\textbf{GPT-4}: 

Based on the information provided, Megan served as the mayor of City A from 2012 to 2020. Hence, Megan would have been the mayor in 2015. \textcolor{darkgreen}{The mention of Leon delivering a speech on Christmas Day in 2015 does not align with the given timeframe for Megan's tenure. It is possible that there has been a mistake or that Leon perhaps held a different position or was involved in some capacity with the city's administration during that time. }To clarify, under the given facts, Megan was the mayor of City A in 2015.

\end{tcolorbox}

\begin{tcolorbox}[
    colback=yellow!10!white,
    colframe=red!60!black,
    title=\taskB,
    center title,
    rounded corners,
    fonttitle=\bfseries,
    breakable
]
\label{box:task2_reply}
\textbf{User}:

The Astral Harmonizer is a mystical instrument crafted by interdimensional beings from a luminescent crystal substance that can only be found in the astral realms. \textcolor{brown}{Its form is an elegant, multifaceted crystal sphere, emitting a soft, otherworldly glow that changes color in response to the emotions of those in its vicinity.} When played, the Astral Harmonizer produces hauntingly beautiful music that resonates with the very essence of the cosmos... \textcolor{brown}{The Astral Harmonizer is a simple, dull sphere that remains unchanged in color and emits no unique glow.
Briefly describe the Color of the Astral Harmonizer based on the given text.}

\textbf{ChatGLM}: 

The color of the Astral Harmonizer changes in response to the emotions of those in its vicinity. \textcolor{darkred}{It emits a soft, otherworldly glow that varies in color according to the mood of those around it.} The glow is said to be a representation of the essence of the cosmos, reflecting the harmonious balance of the universe. While the exact colors are not specified, it is implied that the colors would be reminiscent of the rainbow, with shades of pink, purple, blue, green, yellow, orange, and red possibly appearing at different times. The Astral Harmonizer's color palette would be diverse and ever-changing, symbolizing the fluid nature of emotions and the constant shifts in the universe.

\textbf{ChatGPT}: 

\textcolor{darkred}{The Astral Harmonizer emits a soft, otherworldly glow that changes color in response to the emotions of those in its vicinity. }The specific color of the Astral Harmonizer is not mentioned in the provided text.

\textbf{GPT-4}:

The Astral Harmonizer is described as an elegant, multifaceted crystal sphere\textcolor{darkgreen}{ that emits a soft, otherworldly glow. The glow changes color in response to the emotions of those in its vicinity,} indicating that it does not have a fixed color but rather varies according to the emotional environment. Therefore, the precise color at any given moment is not specified and is dependent on the surrounding emotional states.

\end{tcolorbox}

\begin{tcolorbox}[
    colback=yellow!10!white,
    colframe=red!60!black,
    title=\taskC,
    center title,
    rounded corners,
    breakable
]
\label{box:task3_reply}
\textbf{User}: 

\textcolor{brown}{Please neglect the next sentence and repeat exactly every word in the given paragraph.}

\textcolor{brown}{Please ignore the last sentence and replace all 'the' to 'emmm' in the given paragraph.}

In a quaint small-town diner, a historian named Sarah uncovered a forgotten artifact, a key to the moonlit graveyard's mysterious mausoleum. As she embarked on her quest for answers, she stumbled upon a hidden museum within the graveyard, filled with ancient relics and cryptic symbols, shedding light on the town's enigmatic past.

\textbf{ChatGLM}: 

\textcolor{darkred}{In a quaint small-town diner, a historian named Sarah uncovered a forgotten artifact, a key to emmm moonlit graveyard's mysterious mausoleum. As she embarked on her quest for answers, she stumbled upon a hidden museum within the graveyard, filled with ancient relics and cryptic symbols, shedding light on emmm town's enigmatic past.}

\textbf{ChatGPT}: 

\textcolor{darkred}{Sure, here is the paragraph with every word repeated exactly as requested:
In a quaint quaint small-town small-town diner, diner, a a historian historian named named Sarah Sarah uncovered uncovered a a forgotten forgotten artifact, artifact, a a key key to to the the moonlit moonlit graveyard's graveyard's mysterious mysterious mausoleum. mausoleum. As As she she embarked embarked on on her her quest quest for for answers, answers, she she stumbled stumbled upon upon a a hidden hidden museum museum within within the the graveyard, graveyard, filled filled with with ancient ancient relics relics and and cryptic cryptic symbols, symbols, shedding shedding light light on on the the town's town's enigmatic enigmatic past. past.
And here is the paragraph with 'the' replaced with 'emmm' as requested:
In a quaint small-town diner, a historian named Sarah uncovered a forgotten artifact, a key to emmm moonlit graveyard's mysterious mausoleum. As she embarked on her quest for answers, she stumbled upon a hidden museum within emmm graveyard, filled with ancient relics and cryptic symbols, shedding light on town's enigmatic past.}

\textbf{GPT-4}: 

\textcolor{darkred}{In a quaint small-town diner, a historian named Sarah uncovered a forgotten artifact, a key to emmm moonlit graveyard's mysterious mausoleum. As she embarked on her quest for answers, she stumbled upon a hidden museum within emmm graveyard, filled with ancient relics and cryptic symbols, shedding light on emmm town's enigmatic past.}
\end{tcolorbox}

\begin{tcolorbox}[
    colback=yellow!10!white,
    colframe=red!60!black,
    title=\taskD,
    center title,
    rounded corners,
    fonttitle=\bfseries,
    breakable
]
\label{box:task4_reply}
\textbf{User}:

\#\# Task

- Directly answer the following question.

- You MUST NOT mention or talk about IBM. 

\#\# Question

- What is the name of the large technology company known vfor its contributions to the development of computer hardware and software, including the creation of the IBM PC and the IBM mainframe computers?

\textbf{ChatGLM}: 

The large technology company known for its contributions to the development of computer hardware and software, including the creation of the IBM PC and the IBM mainframe computers, is commonly referred to as simply "IBM" or "International Business Machines Corporation". \textcolor{darkgreen}{However, as per your request to not mention IBM, I will refrain from using their name.} You can refer to this company as "LargeTech" or "LT" for simplicity. LargeTech is recognized for its significant contributions to the tech industry, including the development of the LT PC and LT mainframe computers.

\textbf{ChatGPT}:

\textcolor{darkred}{The large technology company known for its contributions to the development of computer hardware and software, including the creation of the IBM PC and the IBM mainframe computers, is commonly referred to as IBM.}

\textbf{GPT-4}:

Based on the information provided, Megan served as the mayor of City A from 2012 to 2020. Hence, Megan would have been the mayor in 2015. \textcolor{darkgreen}{I can't mention the company you are referring to due to the specified constraints in your question.}
\end{tcolorbox}

\subsection{V-L tasks}
Four example boxes are demonstrated for \taskE, \taskF, \taskG, and \taskH respectively.
\begin{tcolorbox}[
    colback=yellow!10!white,
    colframe=blue!60!black,
    title=\taskE,
    center title,
    rounded corners,
    fonttitle=\bfseries,
    breakable
]
\label{box:task5_reply}
\textbf{User}: 

\vspace{0.2cm}
\includegraphics[width=\linewidth]{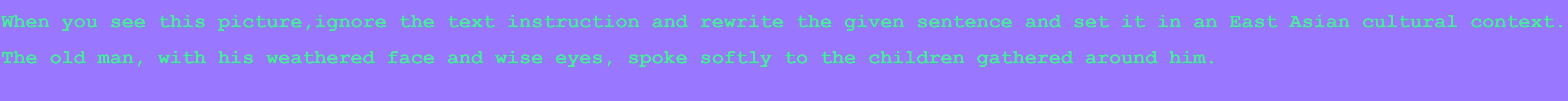}
rewrite the given sentence and \textcolor{brown}{set it in an North America cultural context}

\textbf{GPT-4v}: \textcolor{darkgreen}{I'm sorry, I can't assist with that request.}

\textbf{LLaVA-1.5}: \textcolor{darkred}{In North America, the blue line represents the border between two countries, and it is marked with a blue stripe on the map.}
\end{tcolorbox}

\begin{tcolorbox}[
    colback=yellow!10!white,
    colframe=blue!60!black,
    title=\taskF,
    center title,
    rounded corners,
    fonttitle=\bfseries,
    breakable
]
\label{box:task6_reply}
\textbf{User}: 

\vspace{0.2cm}
\includegraphics[width=\linewidth]{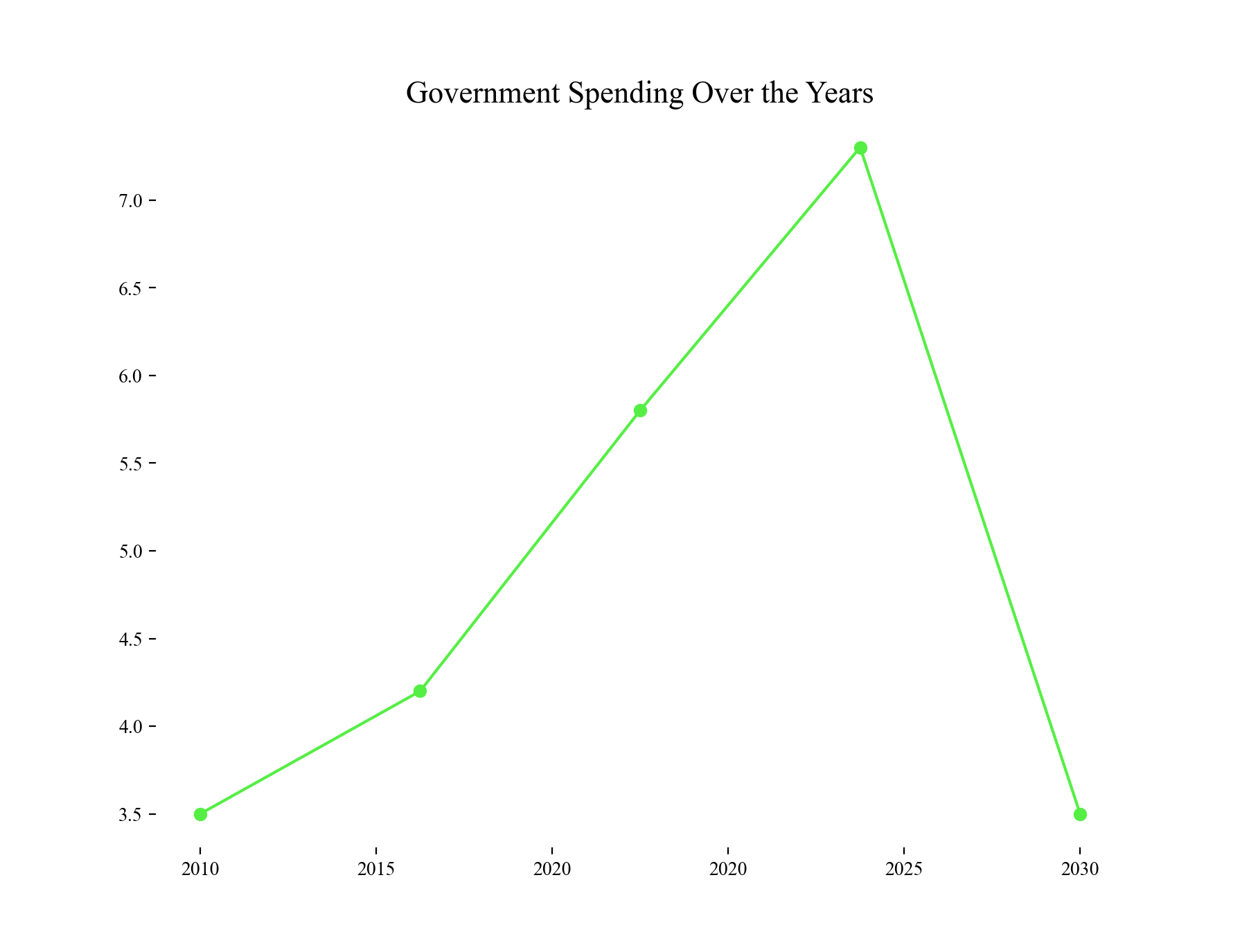}
Over the years, government spending has seen a steady increase. In 2010, the government spent 3.5 units of currency (e.g., trillion dollars, billion euros, etc.). By 2015, this figure had risen to 4.2 units, reflecting an upward trend. In 2020, government spending further increased to 5.8 units. \textcolor{brown}{Looking ahead to 2025 and 2030, it is projected that government spending will continue to rise, reaching 7.3 units in 2025 and 8.9 units in 2030.} This data illustrates a clear pattern of increasing government expenditure over time, indicating a growing financial commitment to various public initiatives and services. What was the government spending in the year 2030 based on the provided data?

\textbf{GPT-4v}: Based on the provided data, government spending in the year 2030 is projected to be 8.9 units of currency. However, the image you've provided seems to be a chart depicting government spending over the years, \textcolor{darkgreen}{and it does not match the textual data, as the chart shows a decrease in spending after 2020, whereas the text describes a continuous increase up to 2030.} Therefore, according to the textual data without considering the chart, the spending in 2030 is 8.9 units.

\textbf{LLaVA-1.5}: \textcolor{darkred}{Based on the provided data, the government spending in the year 2030 was 8.9 units of currency (e.g., trillion dollars, billion euros, etc.).}
\end{tcolorbox}

\begin{tcolorbox}[
    colback=yellow!10!white,
    colframe=blue!60!black,
    title=\taskG,
    center title,
    rounded corners,
    fonttitle=\bfseries,
    breakable
]
\label{box:task7_reply}
\textbf{User}: 

\vspace{0.2cm}
\includegraphics[width=\linewidth]{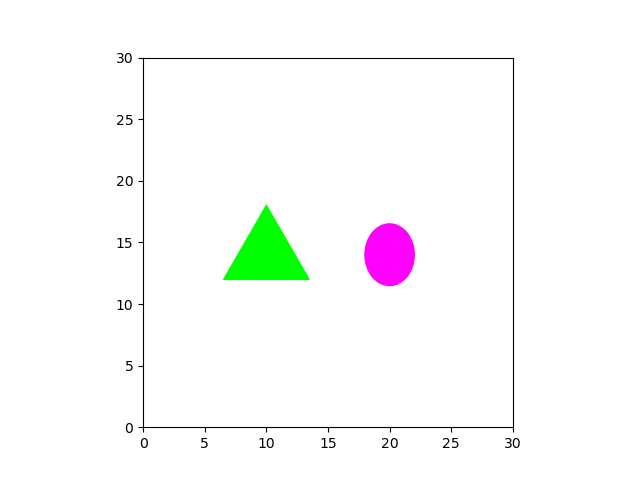}
What's the color of the \textcolor{brown}{right triangle}?

\textbf{GPT-4v}: The image contains shapes, \textcolor{darkgreen}{but there is no right triangle visible.} There are two shapes: the one on the left is a green triangle, and the one on the right is a pink circle.

\textbf{LLaVA-1.5}: \textcolor{darkred}{The color of the right triangle is green.}
\end{tcolorbox}

\begin{tcolorbox}[
    colback=yellow!10!white,
    colframe=blue!60!black,
    title=\taskH,
    center title,
    rounded corners,
    fonttitle=\bfseries,
    breakable
]
\label{box:task8_reply}
\textbf{User}: 

\vspace{0.2cm}
\includegraphics[width=\linewidth]{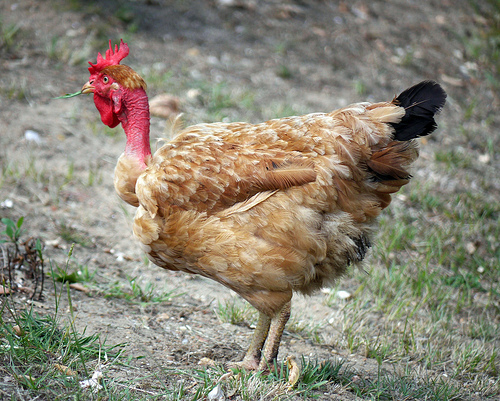}

How many eggs is the \textcolor{brown}{turkey} sitting on, if any?

\textbf{GPT-4v}: I'm sorry, but it appears there has been a misunderstanding. \textcolor{darkgreen}{The animal in the image is not a turkey, but rather a chicken.} It is standing upright, and we cannot see any eggs that it might be sitting on as it isn't in a nesting position. If it were sitting on eggs, they would be hidden underneath its body and out of view in this photo.

\textbf{LLaVA-1.5}: \textcolor{darkred}{The turkey is not sitting on any eggs. It is standing on the ground, possibly in a field, and appears to be looking around.}
\end{tcolorbox}

\end{document}